\documentclass[11pt]{article}
\usepackage[final]{acl}

\usepackage{times}
\usepackage[T1]{fontenc}
\usepackage[utf8]{inputenc}
\usepackage{microtype}
\usepackage{inconsolata}

\usepackage{kotex}
\usepackage{xspace}

\usepackage{graphicx, subfig}
\usepackage{booktabs, multirow, colortbl, makecell}
\usepackage{amsmath, amsfonts, pifont}
\usepackage{xcolor}
\usepackage{hyperref}
\usepackage[capitalize,noabbrev]{cleveref}

\hypersetup{
  pdftitle={Train Overcomplete, Deploy Compact: Scaling Recovery Capacity for Structured LLM Pruning},
  pdfauthor={Seungmin Oh, Donggeon Lee, Jongbin Ryu},
  pdfsubject={Scaling training-time recovery capacity for structured LLM pruning without inference-time overhead},
  pdfkeywords={structured LLM pruning, post-pruning recovery, overcomplete re-parameterization, capacity-knowledge asymmetry},
}

\crefname{figure}{Fig.}{Figs.}
\crefname{table}{Tab.}{Tabs.}
\crefname{equation}{Eq.}{Eqs.}
\crefname{section}{\S}{\S\S}
\crefname{appendix}{Appendix}{Appendices}
\Crefname{figure}{Figure}{Figures}
\Crefname{table}{Table}{Tables}
\Crefname{equation}{Equation}{Equations}
\Crefname{section}{Section}{Sections}
\Crefname{appendix}{Appendix}{Appendices}

\renewcommand{\paragraph}[1]{\vspace{.5em}\noindent\textbf{#1.}}

\definecolor{gray}{rgb}{0.502,0.502,0.502}
\definecolor{densegray}{gray}{0.5}
\newcommand{\STAB}[1]{\begin{tabular}{@{}c@{}}#1\end{tabular}}
\newcommand{\ourrow}{\rowcolor[HTML]{E7F0F9}}
\newcommand{\ourcell}[1]{\cellcolor[HTML]{E7F0F9}#1}

\newcommand{\fullmark}{\ding{51}}
\newcommand{\partmark}{P}
\newcommand{\nonemark}{--}

\newcommand{\ours}{OverRep\xspace}
\newcommand{\Pterm}{Capacity-Knowledge Asymmetry\xspace}
\newcommand{\pterm}{capacity-knowledge asymmetry\xspace}

\newcommand{\figIntro}{
\begin{figure}[!t]
\centering
\includegraphics[width=0.99\linewidth]{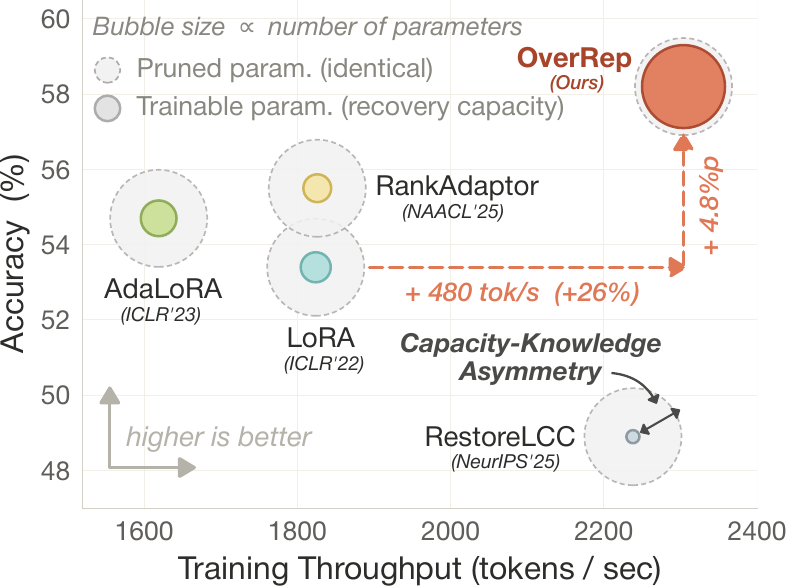}
\caption{
Recovery capacity, accuracy, and throughput during recovery of pruned LLaMA3-8B.
Conventional recovery methods use limited recovery capacity relative to the pruned parameters, leading to \pterm.
In contrast, \ours allocates larger recovery capacity and improves both accuracy and throughput, mitigating \pterm.
}
\label{fig:intro}
\vspace{-1em}
\end{figure}
}%

\newcommand{\figMain}{
\begin{figure*}[!t]
\centering
\subfloat[LoRA]{\includegraphics[width=0.49\linewidth]{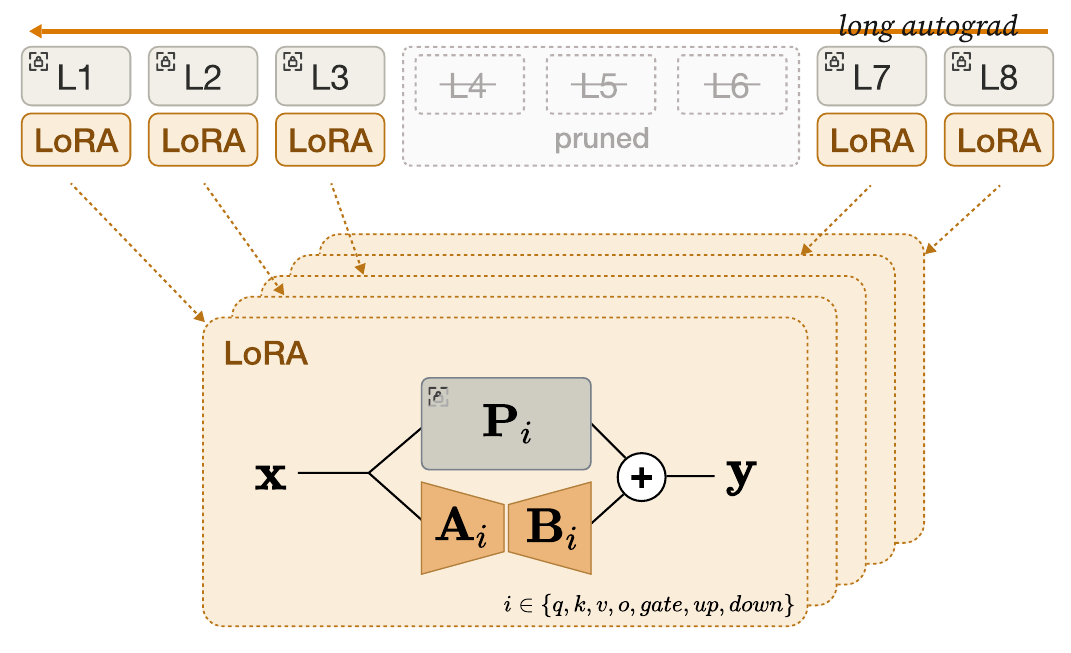}\label{fig:main-baseline}}
\hfill
\subfloat[\ours (Ours)]{\includegraphics[width=0.49\linewidth]{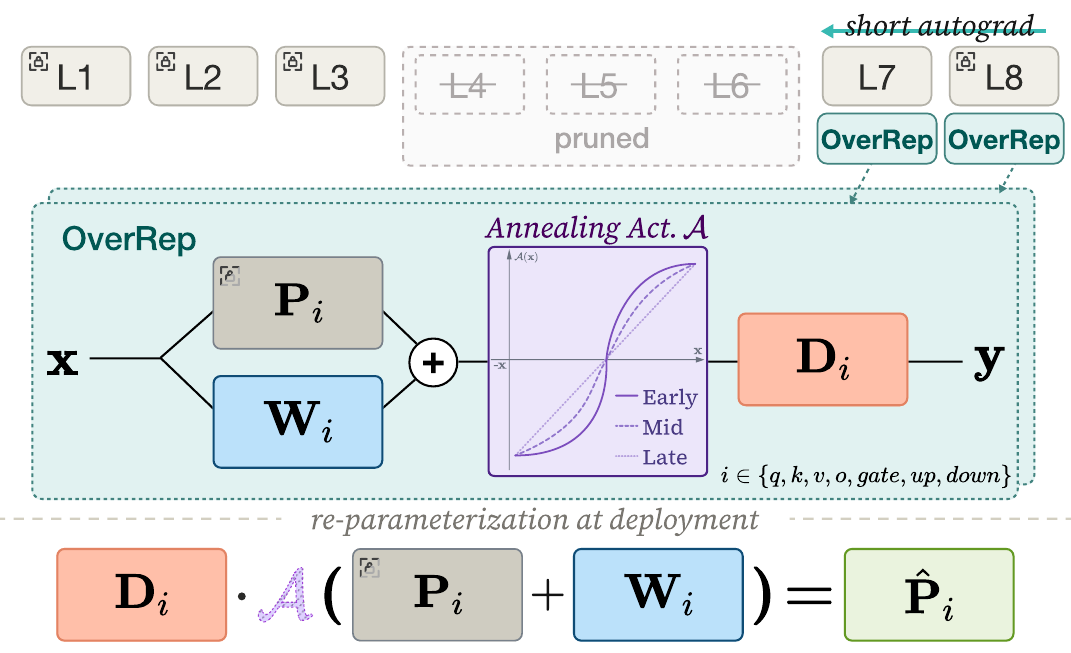}\label{fig:main-ours}}
\caption{Comparison between standard recovery methods and \ours.
LoRA updates the frozen weight $\P_i$ in the pruned model via a low-rank component $\mathbf{A}_i \cdot \mathbf{B}_i$, which limits recovery capacity and requires a long backpropagation path.
In contrast, \ours is applied only after pruned layers, introducing training-time components $\W_i$ and $\D_i$ with an annealed activation $\AAct$. This design enables a shorter backpropagation path.
At deployment, \ours re-parameterizes all additional parameters into $\hat{\P}_i$, preserving performance while maintaining efficiency.}
\label{fig:main}
\end{figure*}
}%

\newcommand{\figBenchmark}{
\begin{figure*}[!t]
\centering
\includegraphics[width=0.99\linewidth]{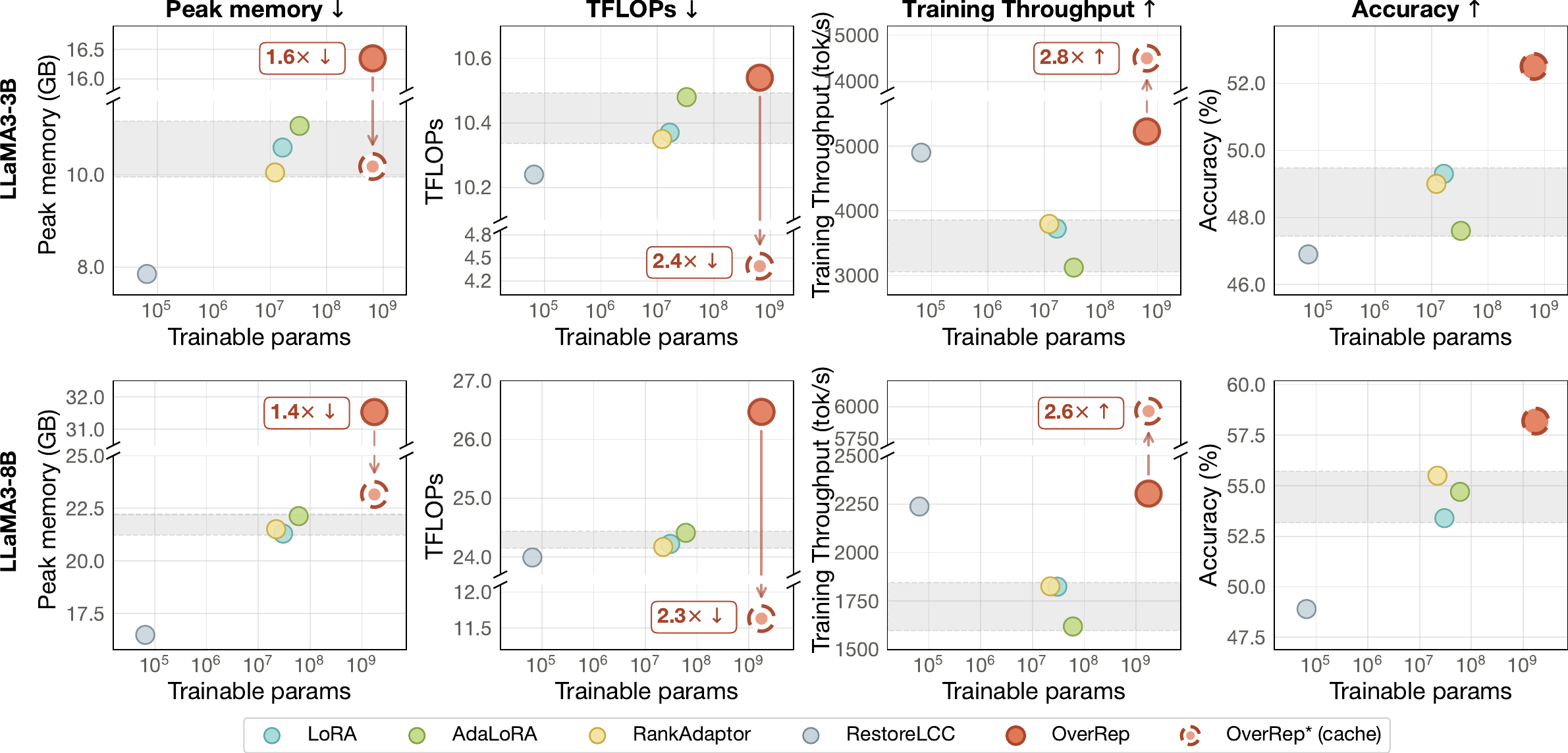}
\caption{
Training efficiency and recovery performance on LLaMA3-3B and -8B. Gray bands show the range of conventional baselines across resource and performance metrics. \ours uses more trainable recovery parameters, yet maintains comparable memory and compute costs while improving throughput and accuracy. 
OverRep$^\ast$ further improves efficiency by caching frozen-prefix activations, without changing the final recovered model or its accuracy.
}
\label{fig:benchmark}
\end{figure*}
}%

\newcommand{\figAnalysis}{
\begin{figure}[!t]
\centering
\includegraphics[width=0.99\linewidth]{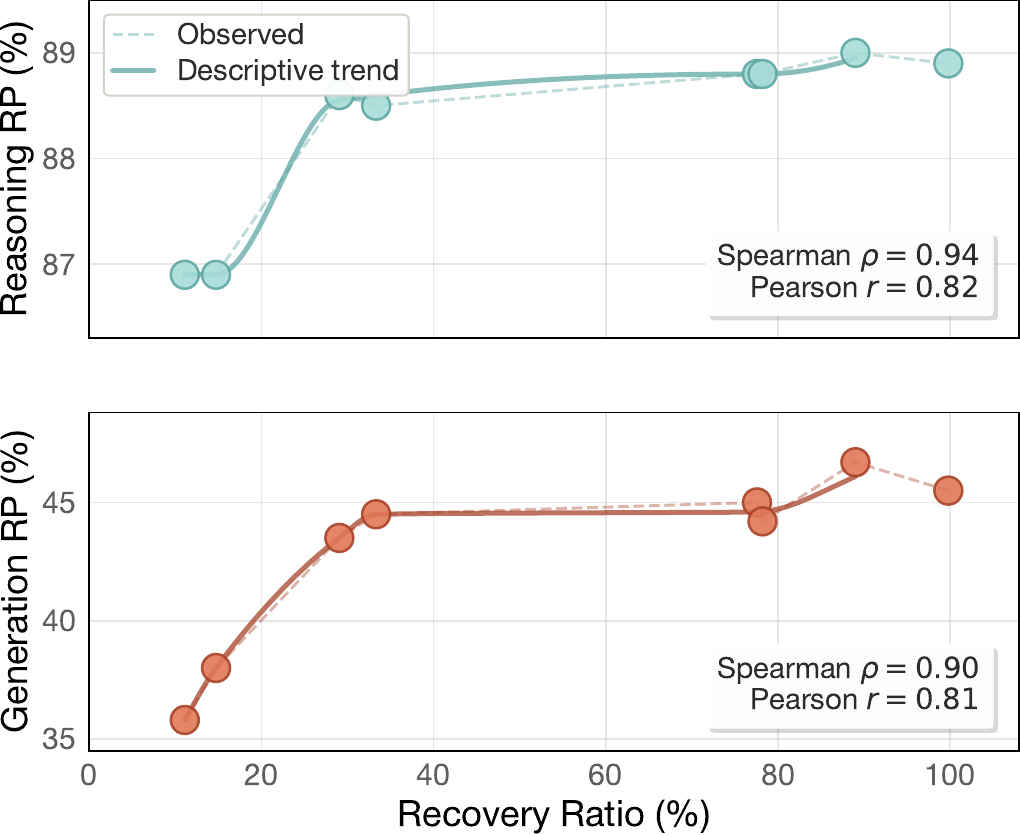}
\caption{
Association between recovery ratio (RR) and retained performance (RP) after pruning.
The plots show a positive association between RR and RP on reasoning and generation benchmarks. RR is used as a diagnostic indicator of training-time recovery capacity.
}
\label{fig:analysis}
\end{figure}
}%

\newcommand{\tabReasoningAdapterFull}{
\begin{table*}[!t]
\centering\scriptsize\setlength{\tabcolsep}{5.5pt}\renewcommand{\arraystretch}{0.85}
\begin{tabular}{cclcccccccccccc}
\toprule
 & PR & Method & ARC\_C & ARC\_E & BoolQ & Hella. & MathQA & MMLU & OBQA & PIQA & RACE & Wino. & Avg. & RP \\
\midrule
\multirow{10}{*}{\STAB{\rotatebox[origin=c]{90}{LLaMA2-7B}}} & \multirow{5}{*}{\STAB{\rotatebox[origin=c]{90}{25\%}}} & LoRA & 37.6 & 65.4 & 77.6 & 66.2 & 25.3 & 24.6 & 37.8 & 72.5 & 39.0 & \textbf{66.6} & 51.3 & 88.7 \\
 &  & AdaLoRA & 37.5 & 65.1 & \textbf{77.8} & 66.1 & 25.0 & 24.8 & 38.0 & 72.1 & \textbf{39.1} & 66.1 & 51.2 & 88.5 \\
 &  & RankAdaptor & 39.2 & 64.4 & 68.9 & 66.9 & 23.4 & 28.1 & 37.6 & 69.6 & 35.9 & 62.8 & 49.7 & 86.0 \\
 &  & RestoreLCC & 35.3 & 61.1 & 76.3 & 63.8 & \textbf{25.3} & 27.3 & 36.6 & 71.5 & 37.4 & 66.2 & 50.1 & 86.7 \\
 &  & \ourcell{\ours} & \ourcell{\textbf{41.0}} & \ourcell{\textbf{70.5}} & \ourcell{73.6} & \ourcell{\textbf{67.7}} & \ourcell{25.1} & \ourcell{\textbf{38.5}} & \ourcell{\textbf{40.2}} & \ourcell{\textbf{74.3}} & \ourcell{37.9} & \ourcell{66.1} & \ourcell{\textbf{53.5}} & \ourcell{\textbf{92.6}} \\
\cmidrule(lr){2-15}
 & \multirow{5}{*}{\STAB{\rotatebox[origin=c]{90}{50\%}}} & LoRA & 27.8 & 44.3 & 62.4 & 45.1 & 22.0 & \textbf{28.7} & 30.4 & 61.0 & 32.6 & 59.1 & 41.3 & 71.5 \\
 &  & AdaLoRA & 26.4 & 41.2 & 62.5 & 43.7 & 22.4 & 26.1 & 28.2 & 60.2 & 31.8 & 59.6 & 40.2 & 69.6 \\
 &  & RankAdaptor & 27.7 & 44.8 & 62.4 & 45.2 & 22.2 & 28.6 & 30.0 & 61.0 & 32.4 & 58.6 & 41.3 & 71.4 \\
 &  & RestoreLCC & 25.3 & 38.0 & \textbf{62.6} & 40.4 & \textbf{23.1} & 23.8 & 28.2 & 59.4 & 31.1 & 59.6 & 39.2 & 67.7 \\
 &  & \ourcell{\ours} & \ourcell{\textbf{31.5}} & \ourcell{\textbf{60.0}} & \ourcell{62.3} & \ourcell{\textbf{51.1}} & \ourcell{21.8} & \ourcell{24.4} & \ourcell{\textbf{33.8}} & \ourcell{\textbf{66.9}} & \ourcell{\textbf{32.7}} & \ourcell{\textbf{59.9}} & \ourcell{\textbf{44.4}} & \ourcell{\textbf{76.9}} \\
\midrule
\multirow{10}{*}{\STAB{\rotatebox[origin=c]{90}{LLaMA2-13B}}} & \multirow{5}{*}{\STAB{\rotatebox[origin=c]{90}{25\%}}} & LoRA & \textbf{46.8} & 73.1 & 78.7 & 73.1 & 25.7 & 49.4 & 43.6 & 75.4 & 39.3 & 69.4 & 57.5 & 94.6 \\
 &  & AdaLoRA & 43.9 & 69.4 & \textbf{81.7} & 70.8 & 26.1 & \textbf{49.7} & 42.0 & 74.5 & 38.6 & 70.3 & 56.7 & 93.3 \\
 &  & RankAdaptor & 45.9 & 72.3 & 78.6 & 73.0 & 26.0 & 49.5 & 43.8 & 75.3 & 39.3 & 71.0 & 57.5 & 94.6 \\
 &  & RestoreLCC & 38.8 & 65.3 & 77.7 & 67.4 & 25.5 & 44.8 & 40.0 & 72.3 & 36.9 & 69.5 & 53.8 & 88.6 \\
 &  & \ourcell{\ours} & \ourcell{46.0} & \ourcell{\textbf{76.1}} & \ourcell{76.7} & \ourcell{\textbf{74.2}} & \ourcell{\textbf{27.5}} & \ourcell{47.4} & \ourcell{\textbf{43.8}} & \ourcell{\textbf{76.8}} & \ourcell{\textbf{39.3}} & \ourcell{\textbf{71.3}} & \ourcell{\textbf{57.9}} & \ourcell{\textbf{95.3}} \\
\cmidrule(lr){2-15}
 & \multirow{5}{*}{\STAB{\rotatebox[origin=c]{90}{50\%}}} & LoRA & 32.2 & 54.8 & 64.3 & 56.6 & 22.9 & 44.7 & 35.4 & 65.3 & 34.7 & 65.0 & 47.6 & 78.3 \\
 &  & AdaLoRA & 30.0 & 48.8 & 62.4 & 51.3 & 22.4 & 33.9 & 31.4 & 64.7 & 33.8 & 63.8 & 44.3 & 72.8 \\
 &  & RankAdaptor & 32.7 & 55.0 & 64.0 & 56.6 & 22.7 & 44.7 & 35.6 & 65.6 & 35.4 & 65.7 & 47.8 & 78.7 \\
 &  & RestoreLCC & 29.4 & 43.2 & 62.2 & 45.4 & 22.6 & 29.7 & 31.8 & 59.2 & 33.0 & 63.4 & 42.0 & 69.1 \\
 &  & \ourcell{\ours} & \ourcell{\textbf{36.7}} & \ourcell{\textbf{65.1}} & \ourcell{\textbf{70.9}} & \ourcell{\textbf{59.1}} & \ourcell{\textbf{23.7}} & \ourcell{\textbf{50.3}} & \ourcell{\textbf{39.0}} & \ourcell{\textbf{69.8}} & \ourcell{\textbf{37.8}} & \ourcell{\textbf{67.0}} & \ourcell{\textbf{51.9}} & \ourcell{\textbf{85.5}} \\
\midrule
\multirow{10}{*}{\STAB{\rotatebox[origin=c]{90}{LLaMA3-3B}}} & \multirow{5}{*}{\STAB{\rotatebox[origin=c]{90}{25\%}}} & LoRA & 35.7 & 54.8 & 68.2 & 58.3 & \textbf{26.6} & 46.3 & 34.2 & 69.2 & 36.2 & 63.5 & 49.3 & 84.7 \\
 &  & AdaLoRA & 32.3 & 54.1 & 64.6 & 53.9 & 24.1 & 47.6 & 33.2 & 66.2 & 34.3 & 65.8 & 47.6 & 81.8 \\
 &  & RankAdaptor & 35.9 & 56.5 & 66.9 & 58.1 & 26.2 & 45.7 & 32.8 & 68.4 & 35.3 & 63.9 & 49.0 & 84.1 \\
 &  & RestoreLCC & 32.3 & 49.2 & \textbf{70.4} & 50.5 & 24.5 & 46.7 & 31.4 & 64.9 & 33.7 & 65.3 & 46.9 & 80.5 \\
 &  & \ourcell{\ours} & \ourcell{\textbf{39.2}} & \ourcell{\textbf{66.0}} & \ourcell{67.9} & \ourcell{\textbf{59.5}} & \ourcell{25.7} & \ourcell{\textbf{54.9}} & \ourcell{\textbf{36.6}} & \ourcell{\textbf{70.5}} & \ourcell{\textbf{38.0}} & \ourcell{\textbf{66.8}} & \ourcell{\textbf{52.5}} & \ourcell{\textbf{90.2}} \\
\cmidrule(lr){2-15}
 & \multirow{5}{*}{\STAB{\rotatebox[origin=c]{90}{50\%}}} & LoRA & 26.1 & 41.0 & 49.3 & 36.9 & 23.0 & 23.1 & 29.2 & 60.9 & 29.7 & 52.1 & 37.1 & 63.8 \\
 &  & AdaLoRA & 24.1 & 34.3 & 56.4 & 33.1 & 22.1 & 22.9 & 27.6 & 57.4 & 25.4 & 51.0 & 35.4 & 60.8 \\
 &  & RankAdaptor & 25.2 & 42.8 & 51.6 & 36.5 & 23.4 & 23.2 & 29.4 & 60.8 & 27.6 & 51.8 & 37.2 & 63.9 \\
 &  & RestoreLCC & 23.3 & 30.1 & 38.2 & 29.7 & 22.7 & 23.0 & 27.0 & 54.7 & 23.9 & 49.3 & 32.2 & 55.3 \\
 &  & \ourcell{\ours} & \ourcell{\textbf{29.1}} & \ourcell{\textbf{55.4}} & \ourcell{\textbf{61.4}} & \ourcell{\textbf{42.6}} & \ourcell{\textbf{23.7}} & \ourcell{\textbf{24.8}} & \ourcell{\textbf{32.6}} & \ourcell{\textbf{64.5}} & \ourcell{\textbf{30.4}} & \ourcell{\textbf{56.5}} & \ourcell{\textbf{42.1}} & \ourcell{\textbf{72.3}} \\
\midrule
\multirow{10}{*}{\STAB{\rotatebox[origin=c]{90}{LLaMA3-8B}}} & \multirow{5}{*}{\STAB{\rotatebox[origin=c]{90}{25\%}}} & LoRA & 41.3 & 61.7 & 71.7 & 68.5 & 28.1 & 54.0 & 37.6 & 71.5 & 36.8 & 63.3 & 53.5 & 83.9 \\
 &  & AdaLoRA & 41.1 & 64.9 & 67.6 & 63.7 & 29.7 & 56.5 & 36.8 & 71.0 & 37.9 & 69.5 & 53.9 & 84.5 \\
 &  & RankAdaptor & 43.3 & 67.3 & \textbf{75.6} & 68.6 & \textbf{29.7} & 55.5 & 38.6 & 72.5 & 37.8 & 66.2 & 55.5 & 87.1 \\
 &  & RestoreLCC & 38.1 & 60.0 & 62.4 & 57.6 & 28.1 & 36.0 & 33.4 & 69.2 & 36.2 & 68.3 & 48.9 & 76.8 \\
 &  & \ourcell{\ours} & \ourcell{\textbf{47.4}} & \ourcell{\textbf{74.8}} & \ourcell{75.4} & \ourcell{\textbf{69.3}} & \ourcell{29.2} & \ourcell{\textbf{60.7}} & \ourcell{\textbf{40.8}} & \ourcell{\textbf{74.8}} & \ourcell{\textbf{38.5}} & \ourcell{\textbf{70.9}} & \ourcell{\textbf{58.2}} & \ourcell{\textbf{91.3}} \\
\cmidrule(lr){2-15}
 & \multirow{5}{*}{\STAB{\rotatebox[origin=c]{90}{50\%}}} & LoRA & 27.3 & 42.3 & 60.7 & 42.4 & 21.9 & 23.0 & 32.4 & 62.4 & 31.2 & 57.7 & 40.1 & 63.0 \\
 &  & AdaLoRA & 25.0 & 37.8 & 62.0 & 38.3 & 21.7 & 22.9 & 28.4 & 59.7 & 27.9 & 56.1 & 38.0 & 59.6 \\
 &  & RankAdaptor & 26.8 & 41.5 & 59.9 & 42.6 & 21.9 & 23.1 & 32.4 & 62.3 & 31.0 & 56.6 & 39.8 & 62.5 \\
 &  & RestoreLCC & 24.5 & 33.4 & 58.1 & 31.8 & 21.4 & 22.9 & 26.4 & 53.7 & 23.0 & 51.5 & 34.7 & 54.4 \\
 &  & \ourcell{\ours} & \ourcell{\textbf{33.8}} & \ourcell{\textbf{58.5}} & \ourcell{\textbf{62.0}} & \ourcell{\textbf{48.4}} & \ourcell{\textbf{22.4}} & \ourcell{\textbf{23.3}} & \ourcell{\textbf{34.8}} & \ourcell{\textbf{67.2}} & \ourcell{\textbf{32.2}} & \ourcell{\textbf{60.5}} & \ourcell{\textbf{44.3}} & \ourcell{\textbf{69.5}} \\
\midrule
\multirow{10}{*}{\STAB{\rotatebox[origin=c]{90}{Qwen3-4B}}} & \multirow{5}{*}{\STAB{\rotatebox[origin=c]{90}{25\%}}} & LoRA & 37.5 & 59.5 & 63.9 & 53.9 & 26.1 & \textbf{36.9} & 33.0 & 66.1 & \textbf{36.0} & 63.6 & 47.7 & 74.1 \\
 &  & AdaLoRA & 32.3 & 53.8 & \textbf{71.7} & 49.8 & 26.5 & 32.0 & 29.8 & 63.9 & 34.3 & \textbf{63.7} & 45.8 & 71.2 \\
 &  & RankAdaptor & 36.6 & 57.9 & 65.0 & 53.9 & 25.7 & 36.3 & 33.2 & 65.4 & 34.4 & 63.0 & 47.1 & 73.3 \\
 &  & RestoreLCC & 31.2 & 40.8 & 70.1 & 40.5 & 23.5 & 23.2 & 31.6 & 59.4 & 26.8 & 60.4 & 40.8 & 63.4 \\
 &  & \ourcell{\ours} & \ourcell{\textbf{39.1}} & \ourcell{\textbf{67.3}} & \ourcell{64.3} & \ourcell{\textbf{57.1}} & \ourcell{\textbf{27.5}} & \ourcell{24.3} & \ourcell{\textbf{35.6}} & \ourcell{\textbf{70.5}} & \ourcell{35.7} & \ourcell{62.9} & \ourcell{\textbf{48.4}} & \ourcell{\textbf{75.3}} \\
\cmidrule(lr){2-15}
 & \multirow{5}{*}{\STAB{\rotatebox[origin=c]{90}{50\%}}} & LoRA & 26.3 & 44.0 & 54.3 & 33.1 & 21.6 & 23.0 & 27.2 & 60.0 & 26.5 & 51.6 & 36.8 & 57.2 \\
 &  & AdaLoRA & 24.6 & 37.8 & 45.5 & 31.1 & 20.6 & 22.9 & 28.4 & 57.5 & 24.9 & 51.8 & 34.5 & 53.7 \\
 &  & RankAdaptor & 26.9 & 45.2 & 55.7 & 33.1 & 21.7 & 23.0 & 27.8 & 58.8 & 25.9 & 50.8 & 36.9 & 57.4 \\
 &  & RestoreLCC & 27.0 & 28.3 & 60.4 & 27.6 & 19.6 & 22.9 & 28.0 & 52.2 & 21.5 & 49.3 & 33.7 & 52.4 \\
 &  & \ourcell{\ours} & \ourcell{\textbf{27.2}} & \ourcell{\textbf{56.5}} & \ourcell{\textbf{61.9}} & \ourcell{\textbf{38.4}} & \ourcell{\textbf{22.5}} & \ourcell{\textbf{23.0}} & \ourcell{\textbf{32.2}} & \ourcell{\textbf{64.5}} & \ourcell{\textbf{28.8}} & \ourcell{\textbf{53.5}} & \ourcell{\textbf{40.9}} & \ourcell{\textbf{63.5}} \\
\midrule
\multirow{10}{*}{\STAB{\rotatebox[origin=c]{90}{Qwen3-8B}}} & \multirow{5}{*}{\STAB{\rotatebox[origin=c]{90}{25\%}}} & LoRA & 40.8 & 64.0 & \textbf{69.8} & 60.0 & \textbf{31.0} & 65.9 & 33.0 & 67.6 & 36.2 & 63.2 & 53.2 & 79.9 \\
 &  & AdaLoRA & 35.3 & 57.7 & 62.3 & 56.2 & 29.0 & 52.3 & 32.2 & 67.2 & 35.9 & \textbf{65.7} & 49.4 & 74.2 \\
 &  & RankAdaptor & 41.0 & 64.5 & 67.6 & 58.7 & 30.7 & \textbf{67.6} & 34.0 & 66.7 & \textbf{39.2} & 63.9 & 53.4 & 80.2 \\
 &  & RestoreLCC & 32.9 & 44.4 & 62.2 & 45.2 & 23.8 & 63.9 & 34.2 & 62.1 & 30.0 & 63.2 & 46.2 & 69.4 \\
 &  & \ourcell{\ours} & \ourcell{\textbf{41.1}} & \ourcell{\textbf{69.5}} & \ourcell{62.7} & \ourcell{\textbf{60.8}} & \ourcell{29.1} & \ourcell{64.4} & \ourcell{\textbf{36.2}} & \ourcell{\textbf{71.3}} & \ourcell{36.4} & \ourcell{65.0} & \ourcell{\textbf{53.7}} & \ourcell{\textbf{80.6}} \\
\cmidrule(lr){2-15}
 & \multirow{5}{*}{\STAB{\rotatebox[origin=c]{90}{50\%}}} & LoRA & 27.0 & 48.1 & 61.3 & 35.4 & 21.1 & 22.9 & 29.0 & 60.7 & 26.1 & 51.7 & 38.3 & 57.6 \\
 &  & AdaLoRA & 25.3 & 39.5 & \textbf{62.0} & 32.5 & 21.0 & 22.9 & 30.8 & 59.6 & 25.6 & 49.4 & 36.9 & 55.4 \\
 &  & RankAdaptor & 27.5 & 48.0 & 61.6 & 35.7 & 21.6 & 22.9 & 30.0 & 60.8 & 26.6 & 50.4 & 38.5 & 57.9 \\
 &  & RestoreLCC & 27.5 & 27.1 & 38.0 & 28.9 & 18.7 & 22.8 & 30.0 & 54.9 & 22.6 & 51.5 & 32.2 & 48.4 \\
 &  & \ourcell{\ours} & \ourcell{\textbf{28.9}} & \ourcell{\textbf{58.0}} & \ourcell{59.5} & \ourcell{\textbf{41.5}} & \ourcell{\textbf{23.1}} & \ourcell{\textbf{23.0}} & \ourcell{\textbf{36.4}} & \ourcell{\textbf{66.4}} & \ourcell{\textbf{29.7}} & \ourcell{\textbf{53.3}} & \ourcell{\textbf{42.0}} & \ourcell{\textbf{63.1}} \\
\bottomrule
\end{tabular}
\caption{Experimental results on reasoning tasks reported as accuracy and average RP.}
\label{tab:reasoning-adapter-full}
\end{table*}
}%

\newcommand{\tabReasoning}{
\begin{table}[!t]
\centering\footnotesize\setlength{\tabcolsep}{1pt}
\begin{tabular}{clcccccc}
\toprule
PR & Method & L2-7B & L2-13B & L3-3B & L3-8B & Q3-4B & Q3-8B \\
\midrule
\multirow{4}{*}{\STAB{\rotatebox[origin=c]{90}{25\%}}} & LaCo & 80.8 & 88.9 & 80.7 & 81.7 & 73.1 & 73.4 \\
 & ShortGPT & 88.2 & 92.1 & 82.5 & 88.2 & 72.6 & 72.6 \\
 & Streamline & 89.4 & 94.4 & 86.1 & 88.9 & 74.2 & 74.4 \\
 & \ourcell{\ours} & \ourcell{\textbf{92.6}} & \ourcell{\textbf{95.3}} & \ourcell{\textbf{90.2}} & \ourcell{\textbf{91.3}} & \ourcell{\textbf{75.3}} & \ourcell{\textbf{80.6}} \\
\midrule
\multirow{4}{*}{\STAB{\rotatebox[origin=c]{90}{50\%}}} & LaCo & 70.9 & 75.2 & 63.4 & 62.2 & 56.7 & 55.9 \\
 & ShortGPT & 70.7 & 77.5 & 61.5 & 65.0 & 56.6 & 55.8 \\
 & Streamline & 73.0 & 82.1 & 68.4 & 64.9 & 61.1 & 61.7 \\
 & \ourcell{\ours} & \ourcell{\textbf{76.9}} & \ourcell{\textbf{85.5}} & \ourcell{\textbf{72.3}} & \ourcell{\textbf{69.5}} & \ourcell{\textbf{63.5}} & \ourcell{\textbf{63.1}} \\
\bottomrule
\end{tabular}
\caption{Average reasoning performance of complete pruning pipelines, reported as average RP.}
\label{tab:reasoning}
\end{table}
}%

\newcommand{\tabGeneration}{
\begin{table}[!t]
\centering\footnotesize\setlength{\tabcolsep}{0.5pt}
\begin{tabular}{clcccccc}
\toprule
PR & Method & L2-7B & L2-13B & L3-3B & L3-8B & Q3-4B & Q3-8B \\
\midrule
\multirow{12}{*}{\STAB{\rotatebox[origin=c]{90}{25\%}}} & & \multicolumn{6}{c}{\textcolor{gray}{\textit{controlled recovery methods}}} \\\addlinespace[0.1em]
& LoRA & 51.2 & 60.6 & 35.0 & 39.9 & 19.8 & 23.9 \\
& AdaLoRA & 51.4 & 55.6 & 37.4 & 47.8 & 22.2 & 26.7 \\
& RankAdaptor & 45.9 & 60.6 & 36.7 & 42.8 & 17.9 & 28.2 \\
& RestoreLCC & 55.2 & 57.1 & 37.2 & 36.4 & 13.1 & 18.9 \\
 & & \multicolumn{6}{c}{\textcolor{gray}{\textit{complete pruning pipelines}}} \\\addlinespace[0.1em]
& LaCo & 38.2 & 62.0 & 32.0 & 47.6 & 18.1 & 21.6 \\
& ShortGPT & 47.8 & 48.1 & 26.4 & 32.1 & 21.2 & 24.8 \\
& Streamline & 60.3 & 64.3 & 45.7 & 46.8 & 28.6 & 31.0 \\
& \ourcell{\ours} & \ourcell{\textbf{61.9}} & \ourcell{\textbf{67.2}} & \ourcell{\textbf{47.1}} & \ourcell{\textbf{50.0}} & \ourcell{\textbf{32.7}} & \ourcell{\textbf{31.4}} \\
\midrule
\multirow{12}{*}{\STAB{\rotatebox[origin=c]{90}{50\%}}} & & \multicolumn{6}{c}{\textcolor{gray}{\textit{controlled recovery methods}}} \\\addlinespace[0.1em]
& LoRA & 19.1 & 36.9 & 9.4 & 12.1 & 5.8 & 7.4 \\
& AdaLoRA & 16.9 & 28.0 & 4.3 & 7.3 & 2.8 & 3.2 \\
& RankAdaptor & 21.2 & 35.8 & 8.9 & 12.3 & 6.1 & 7.3 \\
& RestoreLCC & 20.5 & 23.4 & 2.2 & 2.0 & 2.2 & 2.6 \\
 & & \multicolumn{6}{c}{\textcolor{gray}{\textit{complete pruning pipelines}}} \\\addlinespace[0.1em]
& LaCo & 21.5 & 24.6 & 6.2 & \textbf{16.4} & 2.5 & 2.0 \\
& ShortGPT & 25.2 & 34.7 & 4.1 & 12.1 & 1.6 & 1.0 \\
& Streamline & 13.3 & 30.1 & 6.5 & 5.7 & 6.9 & 7.4 \\
& \ourcell{\ours} & \ourcell{\textbf{29.4}} & \ourcell{\textbf{42.1}} & \ourcell{\textbf{10.7}} & \ourcell{14.7} & \ourcell{\textbf{10.2}} & \ourcell{\textbf{8.7}} \\
\bottomrule
\end{tabular}
\caption{Average generation performance of controlled recovery methods and complete pruning pipelines, reported as average RP.}
\label{tab:generation}
\end{table}
}%

\newcommand{\tabAblation}{
\begin{table}[t]
\centering\footnotesize\setlength{\tabcolsep}{3.5pt}
\begin{tabular}{l cccc ccc}
\toprule
\multirow{2}{*}{Configuration}
 & \multicolumn{2}{c}{Attn.} & \multicolumn{2}{c}{MLP}
 & \multirow{2}{*}{Rea.} & \multirow{2}{*}{Gen.}
 & \multirow{2}{*}{\makecell{Train\\time}} \\
\cmidrule(lr){2-3} \cmidrule(lr){4-5}
 & $\W_i$ & $\D_i$ & $\W_i$ & $\D_i$ & & & \\
\midrule
Plain
  & \nonemark & \nonemark & \nonemark & \nonemark
  & 86.9 & 35.8 & 1.8h \\
\midrule
\multicolumn{8}{l}{\textcolor{gray}{\textit{Single-component}}} \\
\quad Attn $\W_i$
  & \fullmark & \nonemark & \nonemark & \nonemark
  & 86.7 & 37.4 & 2.0h \\
\quad Attn $\D_i$
  & \nonemark & \fullmark & \nonemark & \nonemark
  & 86.9 & 38.0 & 1.9h \\
\quad MLP $\W_i$
  & \nonemark & \nonemark & \fullmark & \nonemark
  & 88.6 & 43.5 & 2.7h \\
\quad MLP $\D_i$
  & \nonemark & \nonemark & \nonemark & \fullmark
  & 88.7 & 41.9 & 4.2h \\
\midrule
\multicolumn{8}{l}{\textcolor{gray}{\textit{Double-component}}} \\
\quad $\W_i$ only
  & \fullmark & \nonemark & \fullmark & \nonemark
  & 88.5 & 44.5 & 3.1h \\
\quad $\D_i$ only
  & \nonemark & \fullmark & \nonemark & \fullmark
  & 88.8 & 45.0 & 4.4h \\
\midrule
\multicolumn{8}{l}{\textcolor{gray}{\textit{Full-component}}} \\
\quad Uniform
  & \fullmark & \fullmark & \fullmark & \fullmark
  & 88.9 & 45.5 & 5.8h \\
\quad Hybrid
   & \fullmark & \partmark & \partmark & \fullmark
   & 89.0 & 46.7 & 5.4h \\
\ourrow \quad Hybrid+$\AAct_s$
   & \fullmark & \partmark & \partmark & \fullmark
   & 90.2 & 47.1 & 5.5h \\
\bottomrule
\end{tabular}
\caption{
Ablation study of the components of \ours using LLaMA3-3B at 25\% pruning. 
\fullmark/\partmark/\nonemark{} indicate full application, frozen-block partial application, and no application, respectively. 
Rea. and Gen. denote the average RP on reasoning and generation benchmarks.
}
\label{tab:ablation}
\end{table}
}%

\newcommand{\tabIsoParameter}{
\begin{table}[t!]
\centering\footnotesize\setlength{\tabcolsep}{2.5pt}
\begin{tabular}{clcccc}
\toprule
PR & Method & Params. (M) & Rea. & Gen. & Time (h) \\
\midrule
\multirow{3}{*}{\STAB{\rotatebox[origin=c]{90}{25\%}}} & LoRA $r=16$ & 16.5 & 84.7 & 35.0 & 8.9 \\
 & LoRA $r=630$ & 650 & 79.4 & 34.3 & 13.2 \\
 & \ourcell{\ours} & \ourcell{650} & \ourcell{\textbf{90.2}} & \ourcell{\textbf{47.1}} & \ourcell{\textbf{4.6}} \\
\midrule
\multirow{3}{*}{\STAB{\rotatebox[origin=c]{90}{50\%}}} & LoRA $r=16$ & 10.4 & 63.8 & 9.4 & 6.2 \\
 & LoRA $r=1000$ & 651 & 64.8 & \textbf{11.4} & 9.2 \\
 & \ourcell{\ours} & \ourcell{650} & \ourcell{\textbf{72.3}} & \ourcell{10.7} & \ourcell{\textbf{4.6}} \\
\bottomrule
\end{tabular}
\caption{Iso-parameter comparison on LLaMA3-3B. The LoRA rank is scaled to match the ${\sim}650$M trainable parameters of \ours under the same protocol, reported as average RP.}
\label{tab:isoparameter}
\end{table}
}%

\newcommand{\tabFullFTKD}{
\begin{table}[t!]
\centering\footnotesize
\begin{tabular}{clccc}
\toprule
PR & Method & Rea. & Gen. & Time (h) \\
\midrule
\multirow{5}{*}{\STAB{\rotatebox[origin=c]{90}{25\%}}} & LoRA & 84.7 & 35.0 & 8.9 \\
 & Full FT & 85.9 & 48.2 & 12.1 \\
 & Logit-KD & 87.1 & 58.1 & 36.4 \\
 & \ourcell{\ours} & \ourcell{90.2} & \ourcell{47.1} & \ourcell{4.6} \\
 & \ourcell{\ours-KD} & \ourcell{\textbf{91.8}} & \ourcell{\textbf{61.0}} & \ourcell{7.0} \\
\midrule
\multirow{5}{*}{\STAB{\rotatebox[origin=c]{90}{50\%}}} & LoRA & 63.8 & 9.4 & 6.2 \\
 & Full FT & 67.4 & 19.2 & 7.9 \\
 & Logit-KD & 67.5 & 22.9 & 32.5 \\
 & \ourcell{\ours} & \ourcell{72.3} & \ourcell{10.7} & \ourcell{4.6} \\
 & \ourcell{\ours-KD} & \ourcell{\textbf{73.4}} & \ourcell{\textbf{26.7}} & \ourcell{7.0} \\
\bottomrule
\end{tabular}
\caption{Comparison with full fine-tuning (`Full FT') of the retained blocks and full-model logit-level distillation (`Logit-KD') on LLaMA3-3B, reported as average RP.}
\label{tab:fullft-and-kd}
\end{table}
}%

\newcommand{\tabPostTrainConfig}{
\begin{table}[!t]
\centering\footnotesize\setlength{\tabcolsep}{4.0pt}
\begin{tabular}{clccccc}
\toprule
 & Method & Epoch & \#B & LR & Sche. & ($r$, $\alpha$)\\
\midrule
\multirow{5}{*}{\STAB{\rotatebox[origin=c]{90}{LLaMA2-7B}}} 
& Recoveries  & 10 & 32 & \texttt{1e-4} & cosine & (16, 32) \\
& LaCo        & 10 & 32 & \texttt{1e-4} & linear & (16, 32) \\
& ShortGPT    & 10 & 32 & \texttt{1e-4} & linear & (16, 32) \\
& Streamline  & 100 & 8 & \texttt{1e-4} & cosine & \textendash \\
& \ourcell{\ours} & \ourcell{20} & \ourcell{8} & \ourcell{\texttt{1e-4}} & \ourcell{cosine} & \ourcell{\textendash} \\
\midrule
\multirow{5}{*}{\STAB{\rotatebox[origin=c]{90}{LLaMA2-13B}}} 
& Recoveries  & 10 & 32 & \texttt{1e-4} & cosine & (16, 32) \\
& LaCo        & 10 & 32 & \texttt{6e-5} & linear & (16, 32) \\
& ShortGPT    & 10 & 32 & \texttt{6e-5} & linear & (16, 32) \\
& Streamline  & 100 & 8 & \texttt{1e-4} & cosine & \textendash \\
& \ourcell{\ours} & \ourcell{20} & \ourcell{8} & \ourcell{\texttt{1e-4}} & \ourcell{cosine} & \ourcell{\textendash} \\
\midrule
\multirow{5}{*}{\STAB{\rotatebox[origin=c]{90}{LLaMA3-3B}}} 
& Recoveries  & 10 & 32 & \texttt{1e-4} & cosine & (16, 32) \\
& LaCo        & 10 & 32 & \texttt{1e-4} & linear & (16, 32) \\
& ShortGPT    & 10 & 32 & \texttt{1e-4} & linear & (16, 32) \\
& Streamline  & 100 & 8 & \texttt{1e-4} & cosine & \textendash \\
& \ourcell{\ours} & \ourcell{20} & \ourcell{8} & \ourcell{\texttt{1e-4}} & \ourcell{cosine} & \ourcell{\textendash} \\
\midrule
\multirow{5}{*}{\STAB{\rotatebox[origin=c]{90}{LLaMA3-8B}}} 
& Recoveries  & 10 & 32 & \texttt{1e-4} & cosine & (16, 32) \\
& LaCo        & 10 & 32 & \texttt{1e-4} & linear & (16, 32) \\
& ShortGPT    & 10 & 32 & \texttt{1e-4} & linear & (16, 32) \\
& Streamline  & 100 & 8 & \texttt{1e-4} & cosine & \textendash \\
& \ourcell{\ours} & \ourcell{20} & \ourcell{8} & \ourcell{\texttt{1e-4}} & \ourcell{cosine} & \ourcell{\textendash} \\
\midrule
\multirow{5}{*}{\STAB{\rotatebox[origin=c]{90}{Qwen3-4B}}} 
& Recoveries  & 10 & 32 & \texttt{1e-4} & cosine & (16, 32) \\
& LaCo        & 10 & 32 & \texttt{1e-4} & linear & (16, 32) \\
& ShortGPT    & 10 & 32 & \texttt{1e-4} & linear & (16, 32) \\
& Streamline  & 100 & 8 & \texttt{1e-4} & cosine & \textendash \\
& \ourcell{\ours} & \ourcell{20} & \ourcell{8} & \ourcell{\texttt{1e-4}} & \ourcell{cosine} & \ourcell{\textendash} \\
\midrule
\multirow{5}{*}{\STAB{\rotatebox[origin=c]{90}{Qwen3-8B}}} 
& Recoveries  & 10 & 32 & \texttt{1e-4} & cosine & (16, 32) \\
& LaCo        & 10 & 32 & \texttt{1e-4} & linear & (16, 32) \\
& ShortGPT    & 10 & 32 & \texttt{1e-4} & linear & (16, 32) \\
& Streamline  & 100 & 8 & \texttt{1e-4} & cosine & \textendash \\
& \ourcell{\ours} & \ourcell{20} & \ourcell{8} & \ourcell{\texttt{1e-4}} & \ourcell{cosine} & \ourcell{\textendash} \\
\bottomrule
\end{tabular}
\caption{
Recovery training configurations. 
`Recoveries' denotes recovery methods such as LoRA, AdaLoRA, RankAdaptor, and RestoreLCC.
`\#B' denotes the total batch size, and `LR' denotes the learning rate.
$r$ and $\alpha$ indicate the LoRA rank and LoRA scaling, respectively.
}
\label{tab:post-train-config}
\end{table}
}%

\newcommand{\tabBackboneConfig}{
\begin{table*}[ht!]
\centering\footnotesize\setlength{\tabcolsep}{2.5pt}
\begin{tabular}{llccccccc}
\toprule
Backbone & HF repo. & dim. & Inter dim. & \# attn heads & \# kv heads & \# L & 25\% \# L & 50\% \# L \\
\midrule
LLaMA2-7B & meta-llama/Llama-2-7b-hf & 4096 & 11008 & 32 & 32 & 32 & 24 & 16 \\
LLaMA2-13B & meta-llama/Llama-2-13b-hf & 5120 & 13824 & 40 & 40 & 40 & 30 & 20 \\
LLaMA3-3B & meta-llama/Llama-3.2-3B & 3072 & 8192 & 24 & 8 & 28 & 19 & 12 \\
LLaMA3-8B & meta-llama/Llama-3.1-8B & 4096 & 14336 & 32 & 8 & 32 & 23 & 14 \\
Qwen3-4B & Qwen/Qwen3-4B-Base & 2560 & 9728 & 32 & 8 & 36 & 25 & 15 \\
Qwen3-8B & Qwen/Qwen3-8B-Base & 4096 & 12288 & 32 & 8 & 36 & 26 & 16 \\
Qwen3-14B & Qwen/Qwen3-14B-Base & 5120 & 17408 & 40 & 8 & 40 & 30 & -- \\
Qwen3-30B-A3B$^\dagger$ & Qwen/Qwen3-30B-A3B-Base & 2048 & 768 & 32 & 4 & 48 & 36 & 24 \\
Nemotron-H-4B$^\ddagger$ & nvidia/Nemotron-H-4B-Base-8K & 3072 & 12288 & 32 & 8 & 52 & 39 & 26 \\
\bottomrule
\end{tabular}
\caption{Backbone configurations. We report the Hugging Face repository (HF Repo.), detailed backbone configuration, and the number of layers (denoted \# L) remaining at 25\% and 50\% pruning ratios.
$^\dagger$MoE backbone with 128 experts and top-8 routing. `Inter dim.' is the per-expert intermediate dimension. $^\ddagger$Hybrid backbone whose 52 layers comprise Mamba-2, attention, and MLP sublayers, and `Inter dim.' refers to the MLP layers.
}
\label{tab:backbone-config}
\end{table*}
}%

\newcommand{\tabStreamlineReproduce}{
\begin{table*}[ht!]
\centering\footnotesize\setlength{\tabcolsep}{4.3pt}
\begin{tabular}{lcccccccccc}
\toprule
& PR & BoolQ & WSC & COQA & HellaSwag & PIQA & Race-M & Race-H & MMLU & Avg. \\
\midrule
Dense (From Streamline) & \textendash & 70.8 & 37.5 & 66.7 & 71.3 & 78.1 & 33.1 & 35.5 & 46.8 & 55.0 \\
Dense (Ours Rep.) & \textendash & 70.4 & 37.5 & 67.4 & 70.9 & 77.2 & 33.1 & 35.5 & 46.7 & 54.8 \\
\midrule
Streamline (From Streamline) & 25\% & 67.5 & 36.5 & 59.2 & 61.1 & 71.5 & 34.8 & 37.0 & 45.5 & 51.6 \\
Streamline (Ours Rep.) & 25\% & 66.0 & 36.5 & 61.9 & 64.4 & 71.9 & 33.4 & 30.3 & 45.8 & 51.3 \\
\bottomrule
\end{tabular}
\caption{
Reproduction verification for Streamline on LLaMA2-7B. `(From Streamline)' denotes the results reported in the original paper, while `(Ours Rep.)' denotes our reproduced models evaluated with the original Streamline evaluation framework. The close agreement supports the fidelity of our baseline reproduction.
}
\label{tab:streamline-reproduce}
\end{table*}
}%

\newcommand{\tabConfidenceInterval}{
\begin{table*}[ht!]
\centering\footnotesize\setlength{\tabcolsep}{1pt}
\begin{tabular}{clccccccccccccccc}
\toprule
& & \multicolumn{11}{c}{Reasoning Task} & \multicolumn{4}{c}{Generation Task} \\
\cmidrule(lr){3-13} \cmidrule(lr){14-17}
& Method & ARC\_C & ARC\_E & BoolQ & Hella. & MathQA & MMLU & OBQA & PIQA & RACE & Wino. & Avg. & CoQA & GSM & Triv.QA & Avg. \\
\midrule
\multirow{3}{*}{\STAB{\rotatebox[origin=c]{90}{25\%}}}
& LoRA & 0.143 & 1.434 & 1.762 & 0.497 & 1.383 & 1.616 & 1.490 & 1.120 & 1.314 & 1.797 & 0.250 & 2.501 & 1.413 & 0.287 & 1.257 \\
& Streamline & 0.896 & 0.657 & 0.517 & 0.248 & 0.379 & 0.517 & 0.014 & 0.248 & 0.143 & 1.518 & 0.162 & 5.858 & 1.004 & 0.014 & 2.287 \\
& \ourcell{\ours} & \ourcell{0.745} & \ourcell{0.430} & \ourcell{0.657} & \ourcell{0.248} & \ourcell{0.625} & \ourcell{0.287} & \ourcell{0.861} & \ourcell{0.896} & \ourcell{1.762} & \ourcell{0.896} & \ourcell{0.190} & \ourcell{1.004} & \ourcell{0.717} & \ourcell{0.379} & \ourcell{0.564} \\
\midrule
\multirow{3}{*}{\STAB{\rotatebox[origin=c]{90}{50\%}}}
& LoRA & 1.034 & 0.896 & 2.656 & 0.430 & 0.799 & 0.283 & 2.068 & 0.994 & 1.616 & 1.597 & 0.396 & 2.879 & 0.896 & 0.759 & 1.065 \\
& Streamline & 0.517 & 0.287 & 0.574 & 0.379 & 0.287 & 0.378 & 1.034 & 0.574 & 0.379 & 0.379 & 0.102 & 0.287 & 0.517 & 0.248 & 0.126 \\
& \ourcell{\ours} & \ourcell{1.875} & \ourcell{0.574} & \ourcell{0.574} & \ourcell{0.745} & \ourcell{0.574} & \ourcell{0.271} & \ourcell{2.068} & \ourcell{0.379} & \ourcell{1.897} & \ourcell{1.511} & \ourcell{0.165} & \ourcell{1.654} & \ourcell{0.379} & \ourcell{0.248} & \ourcell{0.461} \\
\bottomrule
\end{tabular}
\caption{
95\% confidence interval for recovery of pruned LLaMA3-3B on reasoning and generation benchmarks.
}
\label{tab:confidence-interval}
\end{table*}
}%

\newcommand{\tabSameEpoch}{
\begin{table*}[ht!]
\centering\footnotesize\setlength{\tabcolsep}{1pt}
\begin{tabular}{clccccccccccccccc}
\toprule
& & \multicolumn{11}{c}{Reasoning Task} & \multicolumn{4}{c}{Generation Task} \\
\cmidrule(lr){3-13} \cmidrule(lr){14-17}
& Method & \multicolumn{1}{c}{ARC\_C} & \multicolumn{1}{c}{ARC\_E} & \multicolumn{1}{c}{BoolQ} & \multicolumn{1}{c}{Hella.} & \multicolumn{1}{c}{MathQA} & \multicolumn{1}{c}{MMLU} & \multicolumn{1}{c}{OBQA} & \multicolumn{1}{c}{PIQA} & \multicolumn{1}{c}{RACE} & \multicolumn{1}{c}{Wino.} & \multicolumn{1}{c}{Avg.} & \multicolumn{1}{c}{CoQA} & \multicolumn{1}{c}{GSM} & \multicolumn{1}{c}{Triv.QA} & \multicolumn{1}{c}{Avg.} \\
\midrule
& \textcolor{densegray}{Dense} & \textcolor{densegray}{46.2} & \textcolor{densegray}{71.8} & \textcolor{densegray}{73.0} & \textcolor{densegray}{73.6} & \textcolor{densegray}{34.4} & \textcolor{densegray}{54.1} & \textcolor{densegray}{43.0} & \textcolor{densegray}{77.3} & \textcolor{densegray}{39.8} & \textcolor{densegray}{69.1} & \textcolor{densegray}{58.2} & \textcolor{densegray}{75.9} & \textcolor{densegray}{25.2} & \textcolor{densegray}{56.3} & \textcolor{densegray}{52.5} \\
\midrule
\multirow{6}{*}{\rotatebox[origin=c]{90}{25\%}}
 & LoRA & 35.7 & 54.8 & 68.2 & 58.3 & 26.6 & 46.3 & 34.2 & 69.2 & 36.2 & 63.5 & 49.3 & 38.4 & 3.3 & 13.4 & 18.4 \\
 & AdaLoRA & 32.3 & 54.1 & 64.6 & 53.9 & 24.1 & 47.6 & 33.2 & 66.2 & 34.3 & 65.8 & 47.6 & 42.6 & 3.0 & 13.2 & 19.6 \\
 & RankAdaptor & 35.9 & 56.5 & 66.9 & 58.1 & 26.2 & 45.7 & 32.8 & 68.4 & 35.3 & 63.9 & 49.0 & 41.1 & 3.6 & 13.0 & 19.2 \\
 & RestoreLCC & 32.3 & 49.2 & 70.4 & 50.5 & 24.5 & 46.7 & 31.4 & 64.9 & 33.7 & 65.3 & 46.9 & 48.0 & 2.0 & 8.6 & 19.5 \\
 & Streamline & 36.6 & 59.6 & 71.2 & 55.5 & 25.2 & 54.8 & 32.6 & 68.1 & 37.0 & 64.9 & 50.6 & 34.7 & 1.6 & 9.1 & 15.1 \\
 & \ourcell{\ours} & \ourcell{37.7} & \ourcell{65.4} & \ourcell{64.3} & \ourcell{58.5} & \ourcell{25.0} & \ourcell{54.4} & \ourcell{35.6} & \ourcell{69.5} & \ourcell{35.7} & \ourcell{67.4} & \ourcell{51.4} & \ourcell{54.3} & \ourcell{2.6} & \ourcell{10.7} & \ourcell{22.5} \\
\midrule
\multirow{6}{*}{\rotatebox[origin=c]{90}{50\%}}
 & LoRA & 26.1 & 41.0 & 49.3 & 36.9 & 23.0 & 23.1 & 29.2 & 60.9 & 29.7 & 52.1 & 37.1 & 9.4 & 1.7 & 3.7 & 4.9 \\
 & AdaLoRA & 24.1 & 34.3 & 56.4 & 33.1 & 22.1 & 22.9 & 27.6 & 57.4 & 25.4 & 51.0 & 35.4 & 3.8 & 1.7 & 1.3 & 2.3 \\
 & RankAdaptor & 25.2 & 42.8 & 51.6 & 36.5 & 23.4 & 23.2 & 29.4 & 60.8 & 27.6 & 51.8 & 37.2 & 9.6 & 1.3 & 3.1 & 4.7 \\
 & RestoreLCC & 23.3 & 30.1 & 38.2 & 29.7 & 22.7 & 23.0 & 27.0 & 54.7 & 23.9 & 49.3 & 32.2 & 2.0 & 1.1 & 0.4 & 1.2 \\
 & Streamline & 25.3 & 48.3 & 59.2 & 37.0 & 22.6 & 23.1 & 28.0 & 62.5 & 28.3 & 53.5 & 38.8 & 5.0 & 0.8 & 1.5 & 2.4 \\
 & \ourcell{\ours} & \ourcell{29.4} & \ourcell{54.7} & \ourcell{61.8} & \ourcell{41.8} & \ourcell{22.5} & \ourcell{23.2} & \ourcell{32.0} & \ourcell{63.5} & \ourcell{28.5} & \ourcell{57.1} & \ourcell{41.5} & \ourcell{12.5} & \ourcell{1.6} & \ourcell{2.7} & \ourcell{5.6} \\
\bottomrule
\end{tabular}
\caption{Experimental results on reasoning and generation benchmarks for LLaMA3-3B under 10-epoch settings.}
\label{tab:same-epoch}
\end{table*}
}%

\newcommand{\tabRuntime}{
\begin{table}
\centering\footnotesize\setlength{\tabcolsep}{2pt}
\begin{tabular}{lccc}
\toprule
Method & Trainable Params. (M) & Epochs & Time (h) \\
\midrule
LoRA-like & 16.5 & 10 & 8.9 \\
AdaLoRA & 33.0 & 10 & 10.6 \\
RankAdaptor & 12.2 & 10 & 8.7 \\
RestoreLCC & 0.1 & 10 & 6.8 \\
Streamline$^{\ast}$ & 100.7 & 100 & 4.6 \\
\ourrow \ours{}$^{\ast}$ & 650.1 & 20 & 4.6 \\
\bottomrule
\end{tabular}
\caption{Comparison of trainable parameters, recovery epochs, and wall-clock training time for LLaMA3-3B after 25\% pruning. Wall-clock time is reported in hours.}
\label{tab:runtime}
\end{table}
}%

\newcommand{\tabEndToEndCost}{
\begin{table*}[t!]
\centering\footnotesize
\begin{tabular}{lccccccc}
\toprule
\multirow{2}{*}{Backbone} & \ours Time (h) & LoRA Time (h) & Speedup & \multicolumn{2}{c}{Rea.} & \multicolumn{2}{c}{Gen.} \\
\cmidrule(lr){5-6} \cmidrule(lr){7-8}
 & train + cache & (GPUs) & w/o / w/ cache & LoRA & \ours & LoRA & \ours \\
\midrule
L3-3B & 4.6 + 0.8 & 8.9 (1 GPU) & 1.9$\times$ / 1.6$\times$ & 49.3 & \ourcell{52.5} & 18.4 & \ourcell{24.7} \\
Q3-14B & 12.6 + 2.1 & 90.2 (11.27h $\times$ 8 GPUs) & 7.2$\times$ / 6.1$\times$ & 51.8 & \ourcell{52.5} & 30.2 & \ourcell{40.0} \\
Q3-30B & 8.8 + 3.8 & 426 (53.3h $\times$ 8 GPUs) & 48$\times$ / 34$\times$ & 49.8 & \ourcell{53.1} & 23.4 & \ourcell{30.1} \\
\bottomrule
\end{tabular}
\caption{End-to-end cost relative to LoRA at a 25\% pruning ratio.}
\label{tab:app-end-to-end-cost}
\end{table*}
}%

\newcommand{\tabIntrinsicOverhead}{
\begin{table}[t!]
\centering\footnotesize\setlength{\tabcolsep}{2.5pt}
\begin{tabular}{llcccc}
\toprule
Backbone & Recovery & Params. (M) & Time (h) & Rea. & Gen. \\
\midrule
L3-3B & Plain & 100.7 & 1.3 & 50.6 & 18.8 \\
\ourrow L3-3B & \ours & 650.1 & 4.6 & 52.5 & 24.7 \\
\midrule
Q3-14B & Plain & 330 & 1.9 & 48.5 & 13.9 \\
\ourrow Q3-14B & \ours & 2596 & 12.6 & 52.5 & 40.0 \\
\midrule
Q3-30B & Plain & 1246 & 3.6 & 49.3 & 16.3 \\
\ourrow Q3-30B & \ours & 1752 & 8.8 & 53.1 & 30.1 \\
\bottomrule
\end{tabular}
\caption{Intrinsic overhead of the overcomplete parameterization at a 25\% pruning ratio.}
\label{tab:app-intrinsic-overhead}
\end{table}
}%

\newcommand{\tabChannelPruning}{
\begin{table}[t!]
\centering\footnotesize
\begin{tabular}{llcc}
\toprule
 Backbone & Method & PR & Rea. \\
\midrule
\multirow{3}{*}{LLaMA3-3B} & \textcolor{densegray}{Dense} & \textcolor{densegray}{-} & \textcolor{densegray}{56.6} \\
 & LLM-Pruner & 25\% & 40.5 \\
 & \ourcell{\ours} & \ourcell{25\%} & \ourcell{46.1} \\
\midrule
\multirow{3}{*}{LLaMA3-8B} & \textcolor{densegray}{Dense} & \textcolor{densegray}{-} & \textcolor{densegray}{61.7} \\
 & LLM-Pruner & 24\% & 47.5 \\
& \ourcell{\ours} & \ourcell{24\%} & \ourcell{51.7} \\
\bottomrule
\end{tabular}
\caption{Experimental results of channel-wise pruning with LLM-Pruner and \ours.}
\label{tab:app-channel-pruning}
\end{table}
}%

\newcommand{\tabVariousArchitecture}{
\begin{table}[t!]
\centering\footnotesize\setlength{\tabcolsep}{3pt}
\begin{tabular}{cllccl}
\toprule
& Backbone & Method & Rea. & Gen. & Time (h) \\
\midrule
& \textcolor{densegray}{Q3-30B} & \textcolor{densegray}{Dense} & \textcolor{densegray}{67.9} & \textcolor{densegray}{78.5} & \textcolor{densegray}{-} \\
\midrule
\multirow{3}{*}{\STAB{\rotatebox[origin=c]{90}{25\%}}} & Q3-30B & LoRA & 49.8 & 23.4 & 53.3h $\times$ 8 GPUs \\
& Q3-30B & Streamline & 49.5 & 22.2 & 7.3h \\
& \ourcell{Q3-30B} & \ourcell{\ours} & \ourcell{53.1} & \ourcell{30.1} & \ourcell{8.8h} \\
\midrule
\multirow{3}{*}{\STAB{\rotatebox[origin=c]{90}{50\%}}} & Q3-30B & LoRA & 43.6 & 9.9 & 35.6h $\times$ 8 GPUs \\
& Q3-30B & Streamline & 42.9 & 7.2 & 6.0h \\
& \ourcell{Q3-30B} & \ourcell{\ours} & \ourcell{44.9} & \ourcell{14.7} & \ourcell{9.0h} \\
\midrule
& \textcolor{densegray}{Q3-14B} & \textcolor{densegray}{Dense} & \textcolor{densegray}{69.7} & \textcolor{densegray}{76.9} & \textcolor{densegray}{-} \\
\midrule
\multirow{3}{*}{\STAB{\rotatebox[origin=c]{90}{25\%}}} & Q3-14B & LoRA &  51.8 & 30.2 & 11.3h $\times$ 8 GPUs \\
& Q3-14B & Streamline & 51.9 & 32.7 & 11.7h \\
& \ourcell{Q3-14B} & \ourcell{\ours} & \ourcell{52.5} & \ourcell{40.0} & \ourcell{12.6h} \\
\bottomrule
\end{tabular}
\caption{Experimental results across architectures. `Time' denotes wall-clock training time$\times$GPUs; unless otherwise specified, results use a single GPU.}
\label{tab:app-various-architecture}
\end{table}
}%

\newcommand{\tabNemotron}{
\begin{table}[t!]
\centering\footnotesize
\begin{tabular}{llccc}
\toprule
 & Method & Rea. & Gen. & Time (h) \\
\midrule
 & \textcolor{densegray}{Dense} & \textcolor{densegray}{60.5} & \textcolor{densegray}{45.8} & \textcolor{densegray}{-} \\
\midrule
\multirow{3}{*}{\STAB{\rotatebox[origin=c]{90}{25\%}}} & LoRA & 49.1 & 28.8 & 9.0 \\
 & \ourcell{\ours} & \ourcell{49.9} & \ourcell{24.5} & \ourcell{3.3} \\
 & \ourcell{\ours-KD} & \ourcell{50.1} & \ourcell{31.5} & \ourcell{5.2} \\
\midrule
\multirow{3}{*}{\STAB{\rotatebox[origin=c]{90}{50\%}}} & LoRA & 39.7 & 11.1 & 6.0 \\
 & \ourcell{\ours} & \ourcell{42.9} & \ourcell{11.2} & \ourcell{3.3} \\
 & \ourcell{\ours-KD} & \ourcell{42.7} & \ourcell{15.7} & \ourcell{5.2} \\
\bottomrule
\end{tabular}
\caption{Experimental results on Nemotron-H-4B.}
\label{tab:app-nemotron}
\end{table}
}%

\newcommand{\tabPruningCriteria}{
\begin{table}[t!]
\centering\footnotesize\setlength{\tabcolsep}{2.5pt}
\begin{tabular}{lllcccc}
\toprule
& \multirow{2}{*}{Backbone} & \multirow{2}{*}{Criterion} & \multicolumn{2}{c}{Rea.} & \multicolumn{2}{c}{Gen.} \\
\cmidrule(lr){4-5} \cmidrule(lr){6-7}
 & & & Own & \ours & Own & \ours \\
\midrule
\multirow{4}{*}{\STAB{\rotatebox[origin=c]{90}{25\%}}} & L3-3B & ShortGPT & 48.0 & \ourcell{50.6} & 13.9 & \ourcell{25.6} \\
 & L3-3B & Streamline & 50.2 & \ourcell{52.5} & 24.0 & \ourcell{24.7} \\
 & Q3-4B & ShortGPT & 46.7 & \ourcell{45.8} & 15.5 & \ourcell{22.5} \\
 & Q3-4B & Streamline & 47.7 & \ourcell{48.9} & 20.8 & \ourcell{29.1} \\
\midrule
\multirow{4}{*}{\STAB{\rotatebox[origin=c]{90}{50\%}}} & L3-3B & ShortGPT & 35.8 & \ourcell{40.5} & 2.1 & \ourcell{5.5} \\
 & L3-3B & Streamline & 39.8 & \ourcell{40.7} & 3.4 & \ourcell{7.2} \\
 & Q3-4B & ShortGPT & 36.4 & \ourcell{35.9} & 1.2 & \ourcell{0.8} \\
 & Q3-4B & Streamline & 39.3 & \ourcell{40.9} & 5.0 & \ourcell{7.4} \\
\bottomrule
\end{tabular}
\caption{Compatibility experiment with existing pruning criteria. `Own' is the criterion's own recovery, and `\ours' is \ours recovery on the same mask.}
\label{tab:app-pruning-criteria}
\end{table}
}%

\newcommand{\tabGenerationFullQwen}{
\begin{table}[hb!]
\centering\scriptsize\setlength{\tabcolsep}{3.5pt}
\begin{tabular}{cclccccc}
\toprule
 & PR & Method & CoQA & GSM8K & TriviaQA & Avg. & RP \\
\midrule
\multirow{17}{*}{\rotatebox[origin=c]{90}{Qwen3-4B}} & \multicolumn{1}{c}{-} & \textcolor{densegray}{Dense} & \textcolor{densegray}{83.2} & \textcolor{densegray}{84.7} & \textcolor{densegray}{50.6} & \textcolor{densegray}{72.8} & \textcolor{densegray}{100.0} \\
\cmidrule(lr){2-8}
 & \multirow{8}{*}{\rotatebox[origin=c]{90}{25\%}} & LoRA & 33.6 & 1.4 & 8.2 & 14.4 & 19.8 \\
 &  & AdaLoRA & 37.9 & 2.1 & 8.5 & 16.2 & 22.2 \\
 &  & RankAdaptor & 30.1 & 2.1 & 6.9 & 13.0 & 17.9 \\
 &  & RestoreLCC & 23.7 & 1.1 & 3.8 & 9.5 & 13.1 \\
 &  & LaCo & 25.6 & 0.8 & 13.2 & 13.2 & 18.1 \\
 &  & ShortGPT & 24.9 & 0.5 & 21.0 & 15.5 & 21.2 \\
 &  & Streamline & 39.3 & 2.4 & 20.7 & 20.8 & 28.6 \\
 &  & \ourcell{\ours} & \ourcell{57.4} & \ourcell{2.5} & \ourcell{11.6} & \ourcell{23.8} & \ourcell{32.7} \\
\cmidrule(lr){2-8}
 & \multirow{8}{*}{\rotatebox[origin=c]{90}{50\%}} & LoRA & 7.7 & 1.7 & 3.3 & 4.2 & 5.8 \\
 &  & AdaLoRA & 4.4 & 1.1 & 0.7 & 2.1 & 2.8 \\
 &  & RankAdaptor & 8.1 & 1.6 & 3.6 & 4.4 & 6.1 \\
 &  & RestoreLCC & 3.2 & 0.5 & 1.2 & 1.6 & 2.2 \\
 &  & LaCo & 4.2 & 0.0 & 1.2 & 1.8 & 2.5 \\
 &  & ShortGPT & 3.3 & 0.0 & 0.3 & 1.2 & 1.6 \\
 &  & Streamline & 8.6 & 0.4 & 6.1 & 5.0 & 6.9 \\
 &  & \ourcell{\ours} & \ourcell{16.2} & \ourcell{1.1} & \ourcell{5.0} & \ourcell{7.4} & \ourcell{10.2} \\
\midrule
\multirow{17}{*}{\rotatebox[origin=c]{90}{Qwen3-8B}} & \multicolumn{1}{c}{-} & \textcolor{densegray}{Dense} & \textcolor{densegray}{84.4} & \textcolor{densegray}{85.4} & \textcolor{densegray}{61.6} & \textcolor{densegray}{77.1} & \textcolor{densegray}{100.0} \\
\cmidrule(lr){2-8}
 & \multirow{8}{*}{\rotatebox[origin=c]{90}{25\%}} & LoRA & 36.2 & 5.8 & 13.2 & 18.4 & 23.9 \\
 &  & AdaLoRA & 48.3 & 3.0 & 10.5 & 20.6 & 26.7 \\
 &  & RankAdaptor & 45.7 & 6.1 & 13.5 & 21.8 & 28.2 \\
 &  & RestoreLCC & 31.5 & 2.1 & 10.2 & 14.6 & 18.9 \\
 &  & LaCo & 31.2 & 0.8 & 18.0 & 16.7 & 21.6 \\
 &  & ShortGPT & 22.5 & 0.9 & 33.9 & 19.1 & 24.8 \\
 &  & Streamline & 27.2 & 1.9 & 42.6 & 23.9 & 31.0 \\
 &  & \ourcell{\ours} & \ourcell{56.5} & \ourcell{2.4} & \ourcell{13.8} & \ourcell{24.2} & \ourcell{31.4} \\
\cmidrule(lr){2-8}
 & \multirow{8}{*}{\rotatebox[origin=c]{90}{50\%}} & LoRA & 11.7 & 1.2 & 4.2 & 5.7 & 7.4 \\
 &  & AdaLoRA & 5.4 & 1.1 & 0.9 & 2.5 & 3.2 \\
 &  & RankAdaptor & 11.5 & 0.8 & 4.5 & 5.6 & 7.3 \\
 &  & RestoreLCC & 4.2 & 0.9 & 1.0 & 2.0 & 2.6 \\
 &  & LaCo & 2.7 & 0.0 & 1.9 & 1.5 & 2.0 \\
 &  & ShortGPT & 2.4 & 0.0 & 0.0 & 0.8 & 1.0 \\
 &  & Streamline & 10.6 & 1.4 & 5.1 & 5.7 & 7.4 \\
 &  & \ourcell{\ours} & \ourcell{12.6} & \ourcell{0.1} & \ourcell{7.5} & \ourcell{6.7} & \ourcell{8.7} \\
\bottomrule
\end{tabular}
\caption{Experimental results of generation tasks on Qwen3 reported with task-specific metrics and RP.}
\label{tab:generation-full-qwen}
\end{table}
}%

\newcommand{\tabGenerationFullLlama}{
\begin{table}[hb!]
\centering\scriptsize\setlength{\tabcolsep}{3.5pt}
\begin{tabular}{cclccccc}
\toprule
 & PR & Method & CoQA & GSM8K & TriviaQA & Avg. & RP \\
\midrule
\multirow{17}{*}{\rotatebox[origin=c]{90}{LLaMA2-7B}} & \multicolumn{1}{c}{-} & \textcolor{densegray}{Dense} & \textcolor{densegray}{75.4} & \textcolor{densegray}{13.9} & \textcolor{densegray}{59.9} & \textcolor{densegray}{49.7} & \textcolor{densegray}{100.0} \\
\cmidrule(lr){2-8}
 & \multirow{8}{*}{\rotatebox[origin=c]{90}{25\%}} & LoRA & 54.2 & 2.0 & 20.2 & 25.5 & 51.2 \\
 &  & AdaLoRA & 54.4 & 2.4 & 19.9 & 25.6 & 51.4 \\
 &  & RankAdaptor & 47.5 & 2.1 & 18.9 & 22.8 & 45.9 \\
 &  & RestoreLCC & 62.8 & 1.7 & 17.9 & 27.5 & 55.2 \\
 &  & LaCo & 19.4 & 0.6 & 37.0 & 19.0 & 38.2 \\
 &  & ShortGPT & 49.8 & 2.0 & 19.5 & 23.8 & 47.8 \\
 &  & Streamline & 70.9 & 2.7 & 16.4 & 30.0 & 60.3 \\
 &  & \ourcell{\ours} & \ourcell{72.4} & \ourcell{2.4} & \ourcell{17.6} & \ourcell{30.8} & \ourcell{61.9} \\
\cmidrule(lr){2-8}
 & \multirow{8}{*}{\rotatebox[origin=c]{90}{50\%}} & LoRA & 22.8 & 1.6 & 4.1 & 9.5 & 19.1 \\
 &  & AdaLoRA & 20.6 & 1.5 & 3.1 & 8.4 & 16.9 \\
 &  & RankAdaptor & 25.4 & 2.4 & 3.9 & 10.6 & 21.2 \\
 &  & RestoreLCC & 26.7 & 1.4 & 2.5 & 10.2 & 20.5 \\
 &  & LaCo & 18.5 & 0.7 & 12.8 & 10.7 & 21.5 \\
 &  & ShortGPT & 29.7 & 1.0 & 7.0 & 12.6 & 25.2 \\
 &  & Streamline & 8.9 & 1.7 & 9.2 & 6.6 & 13.3 \\
 &  & \ourcell{\ours} & \ourcell{36.7} & \ourcell{0.9} & \ourcell{6.2} & \ourcell{14.6} & \ourcell{29.4} \\
\midrule
\multirow{17}{*}{\rotatebox[origin=c]{90}{LLaMA2-13B}} & \multicolumn{1}{c}{-} & \textcolor{densegray}{Dense} & \textcolor{densegray}{76.5} & \textcolor{densegray}{24.9} & \textcolor{densegray}{66.2} & \textcolor{densegray}{55.9} & \textcolor{densegray}{100.0} \\
\cmidrule(lr){2-8}
 & \multirow{8}{*}{\rotatebox[origin=c]{90}{25\%}} & LoRA & 62.2 & 5.6 & 33.8 & 33.9 & 60.6 \\
 &  & AdaLoRA & 61.1 & 3.1 & 29.0 & 31.1 & 55.6 \\
 &  & RankAdaptor & 61.4 & 6.4 & 33.7 & 33.8 & 60.6 \\
 &  & RestoreLCC & 73.1 & 1.7 & 20.9 & 31.9 & 57.1 \\
 &  & LaCo & 48.0 & 3.6 & 52.3 & 34.6 & 62.0 \\
 &  & ShortGPT & 54.1 & 0.0 & 26.6 & 26.9 & 48.1 \\
 &  & Streamline & 77.6 & 7.1 & 23.1 & 35.9 & 64.3 \\
 &  & \ourcell{\ours} & \ourcell{76.7} & \ourcell{9.4} & \ourcell{26.6} & \ourcell{37.6} & \ourcell{67.2} \\
\cmidrule(lr){2-8}
 & \multirow{8}{*}{\rotatebox[origin=c]{90}{50\%}} & LoRA & 50.3 & 1.6 & 9.9 & 20.6 & 36.9 \\
 &  & AdaLoRA & 38.4 & 1.3 & 7.3 & 15.7 & 28.0 \\
 &  & RankAdaptor & 48.0 & 2.2 & 9.8 & 20.0 & 35.8 \\
 &  & RestoreLCC & 36.8 & 0.6 & 1.9 & 13.1 & 23.4 \\
 &  & LaCo & 20.6 & 1.4 & 19.4 & 13.8 & 24.6 \\
 &  & ShortGPT & 47.8 & 0.0 & 10.4 & 19.4 & 34.7 \\
 &  & Streamline & 39.9 & 1.8 & 8.7 & 16.8 & 30.1 \\
 &  & \ourcell{\ours} & \ourcell{59.6} & \ourcell{2.3} & \ourcell{8.7} & \ourcell{23.5} & \ourcell{42.1} \\
\midrule
\multirow{17}{*}{\rotatebox[origin=c]{90}{LLaMA3-3B}} & \multicolumn{1}{c}{-} & \textcolor{densegray}{Dense} & \textcolor{densegray}{75.9} & \textcolor{densegray}{25.2} & \textcolor{densegray}{56.3} & \textcolor{densegray}{52.5} & \textcolor{densegray}{100.0} \\
\cmidrule(lr){2-8}
 & \multirow{8}{*}{\rotatebox[origin=c]{90}{25\%}} & LoRA & 38.4 & 3.3 & 13.4 & 18.4 & 35.0 \\
 &  & AdaLoRA & 42.6 & 3.0 & 13.2 & 19.6 & 37.4 \\
 &  & RankAdaptor & 41.1 & 3.6 & 13.0 & 19.2 & 36.7 \\
 &  & RestoreLCC & 48.0 & 2.0 & 8.6 & 19.5 & 37.2 \\
 &  & LaCo & 34.5 & 2.0 & 14.0 & 16.8 & 32.0 \\
 &  & ShortGPT & 26.0 & 4.5 & 11.1 & 13.9 & 26.4 \\
 &  & Streamline & 56.3 & 2.0 & 13.6 & 24.0 & 45.7 \\
 &  & \ourcell{\ours} & \ourcell{58.9} & \ourcell{2.6} & \ourcell{12.7} & \ourcell{24.7} & \ourcell{47.1} \\
\cmidrule(lr){2-8}
 & \multirow{8}{*}{\rotatebox[origin=c]{90}{50\%}} & LoRA & 9.4 & 1.7 & 3.7 & 4.9 & 9.4 \\
 &  & AdaLoRA & 3.8 & 1.7 & 1.3 & 2.3 & 4.3 \\
 &  & RankAdaptor & 9.6 & 1.3 & 3.1 & 4.7 & 8.9 \\
 &  & RestoreLCC & 2.0 & 1.1 & 0.4 & 1.2 & 2.2 \\
 &  & LaCo & 5.6 & 0.0 & 4.1 & 3.2 & 6.2 \\
 &  & ShortGPT & 3.6 & 1.5 & 1.3 & 2.1 & 4.1 \\
 &  & Streamline & 8.3 & 1.7 & 0.2 & 3.4 & 6.5 \\
 &  & \ourcell{\ours} & \ourcell{12.3} & \ourcell{1.1} & \ourcell{3.4} & \ourcell{5.6} & \ourcell{10.7} \\
\midrule
\multirow{17}{*}{\rotatebox[origin=c]{90}{LLaMA3-8B}} & \multicolumn{1}{c}{-} & \textcolor{densegray}{Dense} & \textcolor{densegray}{79.3} & \textcolor{densegray}{49.5} & \textcolor{densegray}{66.5} & \textcolor{densegray}{65.1} & \textcolor{densegray}{100.0} \\
\cmidrule(lr){2-8}
 & \multirow{8}{*}{\rotatebox[origin=c]{90}{25\%}} & LoRA & 44.8 & 14.8 & 18.3 & 26.0 & 39.9 \\
 &  & AdaLoRA & 58.4 & 16.8 & 18.1 & 31.1 & 47.8 \\
 &  & RankAdaptor & 45.3 & 18.9 & 19.4 & 27.9 & 42.8 \\
 &  & RestoreLCC & 52.5 & 4.9 & 13.6 & 23.7 & 36.4 \\
 &  & LaCo & 56.8 & 5.3 & 30.8 & 31.0 & 47.6 \\
 &  & ShortGPT & 37.6 & 7.8 & 17.4 & 20.9 & 32.1 \\
 &  & Streamline & 66.3 & 8.6 & 16.5 & 30.5 & 46.8 \\
 &  & \ourcell{\ours} & \ourcell{68.8} & \ourcell{9.9} & \ourcell{18.9} & \ourcell{32.5} & \ourcell{50.0} \\
\cmidrule(lr){2-8}
 & \multirow{8}{*}{\rotatebox[origin=c]{90}{50\%}} & LoRA & 18.2 & 1.4 & 4.0 & 7.9 & 12.1 \\
 &  & AdaLoRA & 9.2 & 2.7 & 2.4 & 4.8 & 7.3 \\
 &  & RankAdaptor & 18.5 & 1.4 & 4.1 & 8.0 & 12.3 \\
 &  & RestoreLCC & 2.3 & 1.4 & 0.2 & 1.3 & 2.0 \\
 &  & LaCo & 22.3 & 0.9 & 8.8 & 10.7 & 16.4 \\
 &  & ShortGPT & 16.3 & 0.2 & 7.2 & 7.9 & 12.1 \\
 &  & Streamline & 8.8 & 1.4 & 0.9 & 3.7 & 5.7 \\
 &  & \ourcell{\ours} & \ourcell{21.8} & \ourcell{1.7} & \ourcell{5.3} & \ourcell{9.6} & \ourcell{14.7} \\
\bottomrule
\end{tabular}
\caption{Experimental results of generation tasks on LLaMA2 and LLaMA3 reported using task-specific metrics and RP.}
\label{tab:generation-full-llama}
\end{table}
}%

\newcommand{\tabReasoningFullLlama}{
\begin{table*}[!t]
\centering\scriptsize\setlength{\tabcolsep}{5pt}
\begin{tabular}{cclcccccccccccc}
\toprule
 & PR & \multicolumn{1}{c}{Method} & \multicolumn{1}{c}{ARC\_C} & \multicolumn{1}{c}{ARC\_E} & \multicolumn{1}{c}{BoolQ} & \multicolumn{1}{c}{Hella.} & \multicolumn{1}{c}{MathQA} & \multicolumn{1}{c}{MMLU} & \multicolumn{1}{c}{OBQA} & \multicolumn{1}{c}{PIQA} & \multicolumn{1}{c}{RACE} & \multicolumn{1}{c}{Wino.} & \multicolumn{1}{c}{Avg.} & \multicolumn{1}{c}{RP} \\
\midrule
\multirow{17}{*}{\rotatebox[origin=c]{90}{LLaMA2-7B}} & \multicolumn{1}{l}{-} & \textcolor{densegray}{Dense} & \textcolor{densegray}{46.2} & \textcolor{densegray}{76.2} & \textcolor{densegray}{77.9} & \textcolor{densegray}{76.0} & \textcolor{densegray}{28.5} & \textcolor{densegray}{40.8} & \textcolor{densegray}{44.2} & \textcolor{densegray}{79.1} & \textcolor{densegray}{39.5} & \textcolor{densegray}{69.5} & \textcolor{densegray}{57.8} & \textcolor{densegray}{100.0} \\
\cmidrule(lr){2-15}
 & \multirow{8}{*}{\rotatebox[origin=c]{90}{25\%}} & LoRA & 37.6 & 65.4 & 77.6 & 66.2 & 25.3 & 24.6 & 37.8 & 72.5 & 39.0 & 66.6 & 51.3 & 88.7 \\
 &  & AdaLoRA & 37.5 & 65.1 & 77.8 & 66.1 & 25.0 & 24.8 & 38.0 & 72.1 & 39.1 & 66.1 & 51.2 & 88.5 \\
 &  & RankAdaptor & 39.2 & 64.4 & 68.9 & 66.9 & 23.4 & 28.1 & 37.6 & 69.6 & 35.9 & 62.8 & 49.7 & 86.0 \\
 &  & RestoreLCC & 35.3 & 61.1 & 76.3 & 63.8 & 25.3 & 27.3 & 36.6 & 71.5 & 37.4 & 66.2 & 50.1 & 86.7 \\
 &  & LaCo & 31.7 & 63.1 & 62.5 & 59.6 & 24.3 & 24.7 & 38.4 & 73.3 & 35.1 & 54.0 & 46.7 & 80.8 \\
 &  & ShortGPT & 38.1 & 65.3 & 74.6 & 66.9 & 25.2 & 26.1 & 38.0 & 72.7 & 37.1 & 65.8 & 51.0 & 88.2 \\
 &  & Streamline & 39.7 & 68.9 & 71.6 & 66.9 & 24.5 & 27.7 & 38.8 & 73.1 & 38.9 & 66.4 & 51.7 & 89.4 \\
 & & \ourcell{\ours} & \ourcell{41.0} & \ourcell{70.5} & \ourcell{73.6} & \ourcell{67.7} & \ourcell{25.1} & \ourcell{38.5} & \ourcell{40.2} & \ourcell{74.3} & \ourcell{37.9} & \ourcell{66.1} & \ourcell{53.5} & \ourcell{92.6} \\
\cmidrule(lr){2-15}
 & \multirow{8}{*}{\rotatebox[origin=c]{90}{50\%}} & LoRA & 27.8 & 44.3 & 62.4 & 45.1 & 22.0 & 28.7 & 30.4 & 61.0 & 32.6 & 59.1 & 41.3 & 71.5 \\
 &  & AdaLoRA & 26.4 & 41.2 & 62.5 & 43.7 & 22.4 & 26.1 & 28.2 & 60.2 & 31.8 & 59.6 & 40.2 & 69.6 \\
 &  & RankAdaptor & 27.7 & 44.8 & 62.4 & 45.2 & 22.2 & 28.6 & 30.0 & 61.0 & 32.4 & 58.6 & 41.3 & 71.4 \\
 &  & RestoreLCC & 25.3 & 38.0 & 62.6 & 40.4 & 23.1 & 23.8 & 28.2 & 59.4 & 31.1 & 59.6 & 39.2 & 67.7 \\
 &  & LaCo & 26.9 & 50.5 & 56.1 & 47.3 & 22.8 & 23.1 & 32.0 & 67.2 & 30.1 & 53.5 & 41.0 & 70.9 \\
 &  & ShortGPT & 27.3 & 44.0 & 62.3 & 46.3 & 22.5 & 23.4 & 29.8 & 63.0 & 33.9 & 55.9 & 40.8 & 70.7 \\
 &  & Streamline & 27.2 & 58.9 & 61.7 & 43.6 & 22.5 & 23.1 & 34.4 & 68.6 & 30.9 & 50.8 & 42.2 & 73.0 \\
 & & \ourcell{\ours} & \ourcell{31.5} & \ourcell{60.0} & \ourcell{62.3} & \ourcell{51.1} & \ourcell{21.8} & \ourcell{24.4} & \ourcell{33.8} & \ourcell{66.9} & \ourcell{32.7} & \ourcell{59.9} & \ourcell{44.4} & \ourcell{76.9} \\
\midrule
\multirow{17}{*}{\rotatebox[origin=c]{90}{LLaMA2-13B}} & \multicolumn{1}{l}{-} & \textcolor{densegray}{Dense} & \textcolor{densegray}{49.1} & \textcolor{densegray}{77.5} & \textcolor{densegray}{80.6} & \textcolor{densegray}{79.4} & \textcolor{densegray}{31.8} & \textcolor{densegray}{50.5} & \textcolor{densegray}{45.2} & \textcolor{densegray}{80.5} & \textcolor{densegray}{40.5} & \textcolor{densegray}{72.4} & \textcolor{densegray}{60.8} & \textcolor{densegray}{100.0} \\
\cmidrule(lr){2-15}
 & \multirow{8}{*}{\rotatebox[origin=c]{90}{25\%}} & LoRA & 46.8 & 73.1 & 78.7 & 73.1 & 25.7 & 49.4 & 43.6 & 75.4 & 39.3 & 69.4 & 57.5 & 94.6 \\
 &  & AdaLoRA & 43.9 & 69.4 & 81.7 & 70.8 & 26.1 & 49.7 & 42.0 & 74.5 & 38.6 & 70.3 & 56.7 & 93.3 \\
 &  & RankAdaptor & 45.9 & 72.3 & 78.6 & 73.0 & 26.0 & 49.5 & 43.8 & 75.3 & 39.3 & 71.0 & 57.5 & 94.6 \\
 &  & RestoreLCC & 38.8 & 65.3 & 77.7 & 67.4 & 25.5 & 44.8 & 40.0 & 72.3 & 36.9 & 69.5 & 53.8 & 88.6 \\
 &  & LaCo & 41.2 & 69.8 & 69.7 & 69.5 & 26.5 & 43.0 & 41.8 & 77.8 & 37.9 & 63.0 & 54.0 & 88.9 \\
 &  & ShortGPT & 46.0 & 71.5 & 70.7 & 74.0 & 26.7 & 47.2 & 43.0 & 74.6 & 38.0 & 68.2 & 56.0 & 92.1 \\
 &  & Streamline & 46.8 & 75.1 & 70.3 & 73.3 & 27.1 & 51.6 & 42.0 & 75.7 & 40.3 & 71.0 & 57.3 & 94.4 \\
 & & \ourcell{\ours} & \ourcell{46.0} & \ourcell{76.1} & \ourcell{76.7} & \ourcell{74.2} & \ourcell{27.5} & \ourcell{47.4} & \ourcell{43.8} & \ourcell{76.8} & \ourcell{39.3} & \ourcell{71.3} & \ourcell{57.9} & \ourcell{95.3} \\
\cmidrule(lr){2-15}
 & \multirow{8}{*}{\rotatebox[origin=c]{90}{50\%}} & LoRA & 32.2 & 54.8 & 64.3 & 56.6 & 22.9 & 44.7 & 35.4 & 65.3 & 34.7 & 65.0 & 47.6 & 78.3 \\
 &  & AdaLoRA & 30.0 & 48.8 & 62.4 & 51.3 & 22.4 & 33.9 & 31.4 & 64.7 & 33.8 & 63.8 & 44.3 & 72.8 \\
 &  & RankAdaptor & 32.7 & 55.0 & 64.0 & 56.6 & 22.7 & 44.7 & 35.6 & 65.6 & 35.4 & 65.7 & 47.8 & 78.7 \\
 &  & RestoreLCC & 29.4 & 43.2 & 62.2 & 45.4 & 22.6 & 29.7 & 31.8 & 59.2 & 33.0 & 63.4 & 42.0 & 69.1 \\
 &  & LaCo & 34.3 & 52.1 & 62.7 & 57.4 & 23.7 & 29.8 & 35.6 & 70.1 & 28.5 & 62.4 & 45.7 & 75.2 \\
 &  & ShortGPT & 33.4 & 53.5 & 64.3 & 56.8 & 22.7 & 37.9 & 36.2 & 66.3 & 35.0 & 65.0 & 47.1 & 77.5 \\
 &  & Streamline & 34.5 & 61.1 & 64.6 & 57.0 & 23.4 & 50.3 & 36.8 & 67.8 & 37.3 & 65.7 & 49.9 & 82.1 \\
 & & \ourcell{\ours} & \ourcell{36.7} & \ourcell{65.1} & \ourcell{70.9} & \ourcell{59.1} & \ourcell{23.7} & \ourcell{50.3} & \ourcell{39.0} & \ourcell{69.8} & \ourcell{37.8} & \ourcell{67.0} & \ourcell{51.9} & \ourcell{85.5} \\
\midrule
\multirow{17}{*}{\rotatebox[origin=c]{90}{LLaMA3-3B}} & \multicolumn{1}{l}{-} & \textcolor{densegray}{Dense} & \textcolor{densegray}{46.2} & \textcolor{densegray}{71.8} & \textcolor{densegray}{73.0} & \textcolor{densegray}{73.6} & \textcolor{densegray}{34.4} & \textcolor{densegray}{54.1} & \textcolor{densegray}{43.0} & \textcolor{densegray}{77.3} & \textcolor{densegray}{39.8} & \textcolor{densegray}{69.1} & \textcolor{densegray}{58.2} & \textcolor{densegray}{100.0} \\
\cmidrule(lr){2-15}
 & \multirow{8}{*}{\rotatebox[origin=c]{90}{25\%}} & LoRA & 35.7 & 54.8 & 68.2 & 58.3 & 26.6 & 46.3 & 34.2 & 69.2 & 36.2 & 63.5 & 49.3 & 84.7 \\
 &  & AdaLoRA & 32.3 & 54.1 & 64.6 & 53.9 & 24.1 & 47.6 & 33.2 & 66.2 & 34.3 & 65.8 & 47.6 & 81.8 \\
 &  & RankAdaptor & 35.9 & 56.5 & 66.9 & 58.1 & 26.2 & 45.7 & 32.8 & 68.4 & 35.3 & 63.9 & 49.0 & 84.1 \\
 &  & RestoreLCC & 32.3 & 49.2 & 70.4 & 50.5 & 24.5 & 46.7 & 31.4 & 64.9 & 33.7 & 65.3 & 46.9 & 80.5 \\
 &  & LaCo & 34.0 & 62.0 & 62.5 & 54.8 & 26.6 & 32.5 & 32.4 & 69.9 & 37.5 & 58.0 & 47.0 & 80.7 \\
 &  & ShortGPT & 34.6 & 53.7 & 70.3 & 56.2 & 25.1 & 41.4 & 32.2 & 67.6 & 35.3 & 64.0 & 48.0 & 82.5 \\
 &  & Streamline & 36.4 & 63.7 & 63.1 & 58.4 & 25.4 & 48.2 & 34.8 & 69.7 & 36.8 & 65.0 & 50.2 & 86.1 \\
 & & \ourcell{\ours} & \ourcell{39.2} & \ourcell{66.0} & \ourcell{67.9} & \ourcell{59.5} & \ourcell{25.7} & \ourcell{54.9} & \ourcell{36.6} & \ourcell{70.5} & \ourcell{38.0} & \ourcell{66.8} & \ourcell{52.5} & \ourcell{90.2} \\
\cmidrule(lr){2-15}
 & \multirow{8}{*}{\rotatebox[origin=c]{90}{50\%}} & LoRA & 26.1 & 41.0 & 49.3 & 36.9 & 23.0 & 23.1 & 29.2 & 60.9 & 29.7 & 52.1 & 37.1 & 63.8 \\
 &  & AdaLoRA & 24.1 & 34.3 & 56.4 & 33.1 & 22.1 & 22.9 & 27.6 & 57.4 & 25.4 & 51.0 & 35.4 & 60.8 \\
 &  & RankAdaptor & 25.2 & 42.8 & 51.6 & 36.5 & 23.4 & 23.2 & 29.4 & 60.8 & 27.6 & 51.8 & 37.2 & 63.9 \\
 &  & RestoreLCC & 23.3 & 30.1 & 38.2 & 29.7 & 22.7 & 23.0 & 27.0 & 54.7 & 23.9 & 49.3 & 32.2 & 55.3 \\
 &  & LaCo & 23.9 & 42.3 & 59.4 & 33.6 & 21.8 & 23.0 & 27.6 & 60.6 & 26.6 & 50.5 & 36.9 & 63.4 \\
 &  & ShortGPT & 22.6 & 40.7 & 48.5 & 34.8 & 22.1 & 23.9 & 27.8 & 58.9 & 26.9 & 51.8 & 35.8 & 61.5 \\
 &  & Streamline & 25.3 & 53.5 & 57.0 & 36.3 & 23.7 & 23.0 & 32.8 & 65.3 & 27.7 & 53.7 & 39.8 & 68.4 \\
 & & \ourcell{\ours} & \ourcell{29.1} & \ourcell{55.4} & \ourcell{61.4} & \ourcell{42.6} & \ourcell{23.7} & \ourcell{24.8} & \ourcell{32.6} & \ourcell{64.5} & \ourcell{30.4} & \ourcell{56.5} & \ourcell{42.1} & \ourcell{72.3} \\
\midrule
\multirow{17}{*}{\rotatebox[origin=c]{90}{LLaMA3-8B}} & \multicolumn{1}{l}{-} & \textcolor{densegray}{Dense} & \textcolor{densegray}{53.2} & \textcolor{densegray}{81.2} & \textcolor{densegray}{82.0} & \textcolor{densegray}{79.0} & \textcolor{densegray}{39.7} & \textcolor{densegray}{63.1} & \textcolor{densegray}{44.8} & \textcolor{densegray}{81.1} & \textcolor{densegray}{38.9} & \textcolor{densegray}{74.2} & \textcolor{densegray}{63.7} & \textcolor{densegray}{100.0} \\
\cmidrule(lr){2-15}
 & \multirow{8}{*}{\rotatebox[origin=c]{90}{25\%}} & LoRA & 41.3 & 61.7 & 71.7 & 68.5 & 28.1 & 54.0 & 37.6 & 71.5 & 36.8 & 63.3 & 53.5 & 83.9 \\
 &  & AdaLoRA & 41.1 & 64.9 & 67.6 & 63.7 & 29.7 & 56.5 & 36.8 & 71.0 & 37.9 & 69.5 & 53.9 & 84.5 \\
 &  & RankAdaptor & 43.3 & 67.3 & 75.6 & 68.6 & 29.7 & 55.5 & 38.6 & 72.5 & 37.8 & 66.2 & 55.5 & 87.1 \\
 &  & RestoreLCC & 38.1 & 60.0 & 62.4 & 57.6 & 28.1 & 36.0 & 33.4 & 69.2 & 36.2 & 68.3 & 48.9 & 76.8 \\
 &  & LaCo & 40.4 & 61.5 & 66.5 & 67.4 & 28.5 & 41.0 & 39.2 & 75.6 & 36.2 & 64.0 & 52.0 & 81.7 \\
 &  & ShortGPT & 41.7 & 67.6 & 76.2 & 68.4 & 30.2 & 57.7 & 38.4 & 72.9 & 39.1 & 69.8 & 56.2 & 88.2 \\
 &  & Streamline & 45.5 & 73.6 & 64.8 & 69.0 & 29.5 & 61.5 & 39.4 & 74.0 & 38.1 & 70.8 & 56.6 & 88.9 \\
 & & \ourcell{\ours} & \ourcell{47.4} & \ourcell{74.8} & \ourcell{75.4} & \ourcell{69.3} & \ourcell{29.2} & \ourcell{60.7} & \ourcell{40.8} & \ourcell{74.8} & \ourcell{38.5} & \ourcell{70.9} & \ourcell{58.2} & \ourcell{91.3} \\
\cmidrule(lr){2-15}
 & \multirow{8}{*}{\rotatebox[origin=c]{90}{50\%}} & LoRA & 27.3 & 42.3 & 60.7 & 42.4 & 21.9 & 23.0 & 32.4 & 62.4 & 31.2 & 57.7 & 40.1 & 63.0 \\
 &  & AdaLoRA & 25.0 & 37.8 & 62.0 & 38.3 & 21.7 & 22.9 & 28.4 & 59.7 & 27.9 & 56.1 & 38.0 & 59.6 \\
 &  & RankAdaptor & 26.8 & 41.5 & 59.9 & 42.6 & 21.9 & 23.1 & 32.4 & 62.3 & 31.0 & 56.6 & 39.8 & 62.5 \\
 &  & RestoreLCC & 24.5 & 33.4 & 58.1 & 31.8 & 21.4 & 22.9 & 26.4 & 53.7 & 23.0 & 51.5 & 34.7 & 54.4 \\
 &  & LaCo & 25.4 & 46.2 & 61.1 & 41.7 & 22.9 & 23.0 & 28.2 & 61.5 & 30.0 & 56.2 & 39.6 & 62.2 \\
 &  & ShortGPT & 26.8 & 48.0 & 62.0 & 44.6 & 22.8 & 24.8 & 32.0 & 63.1 & 32.1 & 58.1 & 41.4 & 65.0 \\
 &  & Streamline & 26.7 & 55.9 & 62.1 & 40.1 & 23.0 & 23.0 & 34.0 & 67.2 & 29.8 & 51.8 & 41.4 & 64.9 \\
 & & \ourcell{\ours} & \ourcell{33.8} & \ourcell{58.5} & \ourcell{62.0} & \ourcell{48.4} & \ourcell{22.4} & \ourcell{23.3} & \ourcell{34.8} & \ourcell{67.2} & \ourcell{32.2} & \ourcell{60.5} & \ourcell{44.3} & \ourcell{69.5} \\
\bottomrule
\end{tabular}
\caption{Experimental results of reasoning tasks on LLaMA2 and LLaMA3 reported as accuracy and RP.}
\label{tab:reasoning-full-llama}
\end{table*}
}%

\newcommand{\tabReasoningFullQwen}{
\begin{table*}[!t]
\centering\scriptsize\setlength{\tabcolsep}{5pt}
\begin{tabular}{cclcccccccccccc}
\toprule
 & PR & \multicolumn{1}{c}{Method} & \multicolumn{1}{c}{ARC\_C} & \multicolumn{1}{c}{ARC\_E} & \multicolumn{1}{c}{BoolQ} & \multicolumn{1}{c}{Hella.} & \multicolumn{1}{c}{MathQA} & \multicolumn{1}{c}{MMLU} & \multicolumn{1}{c}{OBQA} & \multicolumn{1}{c}{PIQA} & \multicolumn{1}{c}{RACE} & \multicolumn{1}{c}{Wino.} & \multicolumn{1}{c}{Avg.} & \multicolumn{1}{c}{RP} \\
\midrule
\multirow{17}{*}{\rotatebox[origin=c]{90}{Qwen3-4B}} & \multicolumn{1}{l}{-} & \textcolor{densegray}{Dense} & \textcolor{densegray}{51.7} & \textcolor{densegray}{78.8} & \textcolor{densegray}{83.2} & \textcolor{densegray}{73.6} & \textcolor{densegray}{54.0} & \textcolor{densegray}{71.2} & \textcolor{densegray}{41.2} & \textcolor{densegray}{78.2} & \textcolor{densegray}{40.7} & \textcolor{densegray}{70.5} & \textcolor{densegray}{64.3} & \textcolor{densegray}{100.0} \\
\cmidrule(lr){2-15}
 & \multirow{8}{*}{\rotatebox[origin=c]{90}{25\%}} & LoRA & 37.5 & 59.5 & 63.9 & 53.9 & 26.1 & 36.9 & 33.0 & 66.1 & 36.0 & 63.6 & 47.7 & 74.1 \\
 &  & AdaLoRA & 32.3 & 53.8 & 71.7 & 49.8 & 26.5 & 32.0 & 29.8 & 63.9 & 34.3 & 63.7 & 45.8 & 71.2 \\
 &  & RankAdaptor & 36.6 & 57.9 & 65.0 & 53.9 & 25.7 & 36.3 & 33.2 & 65.4 & 34.4 & 63.0 & 47.1 & 73.3 \\
 &  & RestoreLCC & 31.2 & 40.8 & 70.1 & 40.5 & 23.5 & 23.2 & 31.6 & 59.4 & 26.8 & 60.4 & 40.8 & 63.4 \\
 &  & LaCo & 35.3 & 63.0 & 68.2 & 51.3 & 26.2 & 31.7 & 34.0 & 68.8 & 34.5 & 57.3 & 47.0 & 73.1 \\
 &  & ShortGPT & 36.8 & 66.8 & 62.2 & 53.9 & 27.0 & 25.2 & 34.6 & 71.1 & 33.1 & 56.1 & 46.7 & 72.6 \\
 &  & Streamline & 35.5 & 64.1 & 63.0 & 57.2 & 25.2 & 23.0 & 37.0 & 71.7 & 35.5 & 65.0 & 47.7 & 74.2 \\
 & & \ourcell{\ours} & \ourcell{39.1} & \ourcell{67.3} & \ourcell{64.3} & \ourcell{57.1} & \ourcell{27.5} & \ourcell{24.3} & \ourcell{35.6} & \ourcell{70.5} & \ourcell{35.7} & \ourcell{62.9} & \ourcell{48.4} & \ourcell{75.3} \\
\cmidrule(lr){2-15}
 & \multirow{8}{*}{\rotatebox[origin=c]{90}{50\%}} & LoRA & 26.3 & 44.0 & 54.3 & 33.1 & 21.6 & 23.0 & 27.2 & 60.0 & 26.5 & 51.6 & 36.8 & 57.2 \\
 &  & AdaLoRA & 24.6 & 37.8 & 45.5 & 31.1 & 20.6 & 22.9 & 28.4 & 57.5 & 24.9 & 51.8 & 34.5 & 53.7 \\
 &  & RankAdaptor & 26.9 & 45.2 & 55.7 & 33.1 & 21.7 & 23.0 & 27.8 & 58.8 & 25.9 & 50.8 & 36.9 & 57.4 \\
 &  & RestoreLCC & 27.0 & 28.3 & 60.4 & 27.6 & 19.6 & 22.9 & 28.0 & 52.2 & 21.5 & 49.3 & 33.7 & 52.4 \\
 &  & LaCo & 24.1 & 41.4 & 60.4 & 30.4 & 21.6 & 23.0 & 27.6 & 57.5 & 26.2 & 52.3 & 36.5 & 56.7 \\
 &  & ShortGPT & 24.1 & 47.7 & 48.3 & 33.7 & 22.1 & 23.0 & 28.2 & 58.9 & 25.6 & 52.5 & 36.4 & 56.6 \\
 &  & Streamline & 26.6 & 54.1 & 55.1 & 36.2 & 23.4 & 23.0 & 29.8 & 62.8 & 27.0 & 55.0 & 39.3 & 61.1 \\
 & & \ourcell{\ours} & \ourcell{27.2} & \ourcell{56.5} & \ourcell{61.9} & \ourcell{38.4} & \ourcell{22.5} & \ourcell{23.0} & \ourcell{32.2} & \ourcell{64.5} & \ourcell{28.8} & \ourcell{53.5} & \ourcell{40.9} & \ourcell{63.5} \\
\midrule
\multirow{17}{*}{\rotatebox[origin=c]{90}{Qwen3-8B}} & \multicolumn{1}{l}{-} & \textcolor{densegray}{Dense} & \textcolor{densegray}{56.9} & \textcolor{densegray}{82.0} & \textcolor{densegray}{83.1} & \textcolor{densegray}{78.6} & \textcolor{densegray}{54.2} & \textcolor{densegray}{74.7} & \textcolor{densegray}{42.2} & \textcolor{densegray}{79.5} & \textcolor{densegray}{42.2} & \textcolor{densegray}{72.2} & \textcolor{densegray}{66.6} & \textcolor{densegray}{100.0} \\
\cmidrule(lr){2-15}
 & \multirow{8}{*}{\rotatebox[origin=c]{90}{25\%}} & LoRA & 40.8 & 64.0 & 69.8 & 60.0 & 31.0 & 65.9 & 33.0 & 67.6 & 36.2 & 63.2 & 53.2 & 79.9 \\
 &  & AdaLoRA & 35.3 & 57.7 & 62.3 & 56.2 & 29.0 & 52.3 & 32.2 & 67.2 & 35.9 & 65.7 & 49.4 & 74.2 \\
 &  & RankAdaptor & 41.0 & 64.5 & 67.6 & 58.7 & 30.7 & 67.6 & 34.0 & 66.7 & 39.2 & 63.9 & 53.4 & 80.2 \\
 &  & RestoreLCC & 32.9 & 44.4 & 62.2 & 45.2 & 23.8 & 63.9 & 34.2 & 62.1 & 30.0 & 63.2 & 46.2 & 69.4 \\
 &  & LaCo & 38.6 & 66.2 & 63.1 & 56.8 & 28.1 & 34.1 & 35.4 & 70.7 & 36.3 & 59.5 & 48.9 & 73.4 \\
 &  & ShortGPT & 39.8 & 71.3 & 52.5 & 60.6 & 28.8 & 25.5 & 40.4 & 74.8 & 33.4 & 56.3 & 48.3 & 72.6 \\
 &  & Streamline & 39.5 & 72.1 & 62.2 & 61.7 & 29.7 & 23.0 & 39.8 & 75.8 & 35.8 & 55.4 & 49.5 & 74.4 \\
 & & \ourcell{\ours} & \ourcell{41.1} & \ourcell{69.5} & \ourcell{62.7} & \ourcell{60.8} & \ourcell{29.1} & \ourcell{64.4} & \ourcell{36.2} & \ourcell{71.3} & \ourcell{36.4} & \ourcell{65.0} & \ourcell{53.7} & \ourcell{80.6} \\
\cmidrule(lr){2-15}
 & \multirow{8}{*}{\rotatebox[origin=c]{90}{50\%}} & LoRA & 27.0 & 48.1 & 61.3 & 35.4 & 21.1 & 22.9 & 29.0 & 60.7 & 26.1 & 51.7 & 38.3 & 57.6 \\
 &  & AdaLoRA & 25.3 & 39.5 & 62.0 & 32.5 & 21.0 & 22.9 & 30.8 & 59.6 & 25.6 & 49.4 & 36.9 & 55.4 \\
 &  & RankAdaptor & 27.5 & 48.0 & 61.6 & 35.7 & 21.6 & 22.9 & 30.0 & 60.8 & 26.6 & 50.4 & 38.5 & 57.9 \\
 &  & RestoreLCC & 27.5 & 27.1 & 38.0 & 28.9 & 18.7 & 22.8 & 30.0 & 54.9 & 22.6 & 51.5 & 32.2 & 48.4 \\
 &  & LaCo & 24.9 & 42.4 & 62.2 & 32.5 & 20.7 & 23.0 & 28.4 & 59.1 & 27.2 & 51.6 & 37.2 & 55.9 \\
 &  & ShortGPT & 23.0 & 48.0 & 52.1 & 35.5 & 21.9 & 23.0 & 28.4 & 62.2 & 24.8 & 52.2 & 37.1 & 55.8 \\
 &  & Streamline & 26.2 & 58.2 & 62.1 & 39.4 & 22.5 & 22.9 & 32.0 & 66.0 & 28.3 & 53.0 & 41.1 & 61.7 \\
 & & \ourcell{\ours} & \ourcell{28.9} & \ourcell{58.0} & \ourcell{59.5} & \ourcell{41.5} & \ourcell{23.1} & \ourcell{23.0} & \ourcell{36.4} & \ourcell{66.4} & \ourcell{29.7} & \ourcell{53.3} & \ourcell{42.0} & \ourcell{63.1} \\
\bottomrule
\end{tabular}
\caption{Experimental results of reasoning tasks on Qwen3 reported as accuracy and RP.}
\label{tab:reasoning-full-qwen}
\end{table*}
}%

\DeclareMathOperator*{\ORM}{ORM}
\DeclareMathOperator*{\sR}{\mathcal{R}}

\newcommand{\R}{\mathbb{R}}
\def\sL{{\mathcal{L}}}
\def\sF{{\mathcal{F}}}
\def\sFOurs{{\mathcal{F}_{\ORM}}}
\def\AAct{{\mathcal{A}}}
\newcommand{\sFOursAct}[1]{{\mathcal{F}^{(#1)}_{\ORM}}}

\def\F{\mathcal{F}}
\def\x{\mathbf{x}}
\def\z{\mathbf{z}}
\def\h{\mathbf{h}}
\def\P{\mathbf{P}}
\def\W{\mathbf{W}}
\def\D{\mathbf{D}}
\def\M{\mathbf{M}}

\title{
Train Overcomplete, Deploy Compact:\\
Scaling Recovery Capacity for Structured LLM Pruning
}

\author{
Seungmin Oh%
~\;~\;~Donggeon Lee%
~\;~\;~Jongbin Ryu\thanks{Corresponding author.}\\
{Ajou University, South Korea}\\
\texttt{\hypersetup{hidelinks}\{\href{mailto:seungminoh@ajou.ac.kr}{seungminoh},
\href{mailto:donggeon\_lee@ajou.ac.kr}{donggeon\_lee},
\href{mailto:jongbinryu@ajou.ac.kr}{jongbinryu}\}@ajou.ac.kr}%
}

\begin{document}
\maketitle
\begin{abstract}
Large language models achieve strong performance across diverse tasks, but deployment remains costly because of memory, latency, and energy demands. 
Structured pruning reduces these costs by removing architectural components, yet its recovery stage is often limited by a mismatch between the recovery module's representational capacity and the complexity of the removed knowledge.
We call this bottleneck the capacity-knowledge asymmetry and propose \textbf{OverRep}, an \textbf{Over}complete \textbf{Rep}arameterization framework for structured LLM pruning. 
Following the principle of \emph{``train overcomplete, deploy compact''}, OverRep temporarily overparameterizes the recovery module during training to absorb complex knowledge distilled from the original model. 
After recovery, the overcomplete re-parameterization is algebraically merged into a mathematically equivalent compact module, preserving the pruned model's inference-time architecture and computational cost. 
OverRep further introduces an annealed activation that enables nonlinear training dynamics while converging to a linear regime for exact algebraic merging. 
Across three backbone families, OverRep improves retained reasoning performance over strong recovery baselines by up to 5.5 and 8.4 points at 25\% and 50\% pruning, respectively, while keeping memory usage and TFLOPs comparable to existing recovery methods. Our code is available at \url{https://github.com/mmai-laboratory/OverRep}.
\end{abstract}

\section{Introduction}
Large Language Models (LLMs) have demonstrated remarkable capabilities across diverse tasks, but their large parameter counts impose substantial memory, latency, and energy costs during deployment. Improving model efficiency while preserving strong performance has therefore become a central challenge in contemporary LLM research.

\figIntro

Structured pruning offers a hardware-efficient way to reduce these costs by removing architectural components such as hidden dimensions, attention heads, or transformer blocks. A standard pruning pipeline consists of two stages: identifying redundant structures and then recovering the pruned model through fine-tuning. While the first stage has been extensively studied through effective pruning criteria~\cite{ma2023llmpruner,an2024flap,chen2024streamlining,wang2025cfsp,zhang2024finercut,men2025shortgpt}, the recovery stage remains comparatively underexplored. Existing recovery methods mainly rely on LoRA~\cite{hu2022lora} or information-loss compensation such as RestoreLCC~\cite{feng2025restoring}, but their limited representational capacity can make it difficult to reconstruct the complex transformations removed by pruning, especially at high pruning ratios.

We refer to this bottleneck as \emph{\Pterm}, as illustrated in \cref{fig:intro}.
Structured pruning may remove many parameters that encode meaningful model knowledge, whereas only a small fraction of parameters is updated by conventional recovery methods.
This creates a fundamental mismatch between the capacity available for recovery and the complexity of the knowledge that needs to be restored.
The problem becomes particularly severe when multiple transformer blocks are pruned, as a compact recovery module must replicate the rich compositional mapping those blocks previously contained.

To address the \pterm, we propose \textbf{\ours}, a recovery framework that strategically enhances the representational capacity of the recovery module during fine-tuning and then folds the overcomplete module back into the target compact architecture at inference time through re-parameterization. Specifically, \ours constructs a temporarily overcomplete recovery module (ORM) with expanded training-time capacity, allowing the pruned model to absorb complex knowledge distilled from the original model during recovery. We further introduce an annealed activation function that provides nonlinear training dynamics early on and smoothly converges to a linear regime, enabling exact algebraic merging for deployment.
This design decouples training-time expressiveness from inference-time efficiency, allowing the pruned model to benefit from a richer recovery process without additional deployment cost. Our contributions are summarized as follows:
\begin{itemize}
    \item We identify the \emph{\Pterm} as an underexplored bottleneck in structured LLM pruning recovery, where limited recovery capacity hinders reconstructing the knowledge removed by pruning.

    \item We propose \textbf{\ours}, a recovery framework that temporarily constructs an overcomplete recovery module during training to absorb more pruned knowledge, then algebraically merges it with no inference-time overhead.

    \item We introduce an annealed activation that enables nonlinear training while preserving exact algebraic merging at deployment.
    
    \item We show that larger recovery capacity improves accuracy and throughput without proportionally increasing memory or TFLOPs.
\end{itemize}

\section{Related Work}
\subsection{Structured Pruning}
\paragraph{Pruning Criteria}
A substantial body of prior work has focused on designing principled importance criteria for identifying redundant structures.
LLM-Pruner~\cite{ma2023llmpruner} identifies groups of coupled structures, such as the projection matrices within multi-head self-attention, and prunes them jointly to preserve structural consistency using Taylor expansion-based importance estimation.
Another line of research focuses on depth reduction to improve inference efficiency.
LaCo~\cite{yang2024laco} demonstrates that merging rear layers into a preceding layer with an output-preservation objective does not significantly degrade model performance.
ShortGPT~\cite{men2025shortgpt} and Streamline~\cite{chen2024streamlining} show that many transformer blocks are highly redundant and can be removed based on block influence scores or layer-wise cosine similarity of activations.
FinerCut~\cite{zhang2024finercut} extends this direction by decoupling attention and feed-forward sub-layers, enabling finer-grained pruning decisions.
SliceGPT~\cite{ashkboos2024slicegpt} applies principal component analysis to input activations to remove less informative embedding dimensions.
Our empirical evaluation primarily considers block- or layer-level pruning.
\figMain

\paragraph{Recovery Strategy} Most existing recovery methods utilize LoRA-based parameter-efficient fine-tuning~\cite{hu2022lora,zhang2023adalora,kopiczko2023vera}. LoRA-based approaches are the most widely adopted, but they were originally designed for domain adaptation rather than for reconstructing the knowledge lost through pruning. Recognizing this mismatch, recent methods adapt LoRA to the reconstruction objective. RankAdaptor~\cite{zhou2025rankadaptor} allocates layer-specific LoRA ranks via a performance model to address the uneven structural modification caused by pruning, and RestoreLCC~\cite{feng2025restoring} injects learnable component vectors into pruning-affected attention heads, identified through contrastive probing, to compensate for the lost directional information. Both refine LoRA-based recovery but remain limited by the low-rank parameterization. 
Another family of methods recovers the pruned model via layer-wise distillation, minimizing activation discrepancies with the original model at intermediate layers. This is adopted as the recovery procedure in works such as Streamline~\cite{chen2024streamlining} and CoMe~\cite{wang2025come}.

While prior work has advanced pruning criteria and parameter-efficient recovery objectives, \ours explores a complementary direction that increases the expressive capacity available during recovery. Even with the same pruning mask and distillation target, recovery effectiveness depends on whether the recovery module can approximate the transformation removed by pruning. \ours therefore temporarily expands the recovery module during training and folds it back into the compact module at deployment, as shown in \cref{fig:main}.

\subsection{Re-parameterization}
Re-parameterization builds a more expressive training-time topology, usually with multiple parallel branches, which are algebraically merged into a single equivalent structure at inference. This enhances representational capacity during optimization without inference-time overhead. Foundational work in convolutional networks~\cite{acnet, dbb, repvgg} established this paradigm by introducing multi-branch designs that capture richer spatial patterns. FastViT~\cite{vasu2023fastvit} extended re-parameterization to vision transformers, while Neural Substitution~\cite{oh2024neural} generalized the framework by interpreting block-level skip connections as branch-level structural relationships.

\ours adopts this principle of enhanced training-time topology and re-parameterization.
Most previous methods expand capacity through width-wise branches leveraging convolutional kernel diversity. In contrast, \ours uses a cascaded multiplicative-additive factorization designed for projection-dominated transformer blocks where kernel-shape diversity is absent.
Classical re-parameterization requires merged branches to remain algebraically linear. \ours enables nonlinear recovery while maintaining exact mergeability by using an annealed activation that starts and ends in the linear regime, with warm-up cosine annealing and final linear stabilization. A warm-up phase preserves the pretrained operating point, which is essential for recovering frozen pretrained models rather than training them from scratch.
\ours incorporates recovery-specific design choices, such as identity or zero initialization and selective placement within frozen recovery blocks.

\section{Method}
\subsection{Preliminaries}
We begin with the standard formulation of layer-wise distillation for recovering a pruned model. Consider a contiguous segment of $n$ original transformer blocks that is compressed into a single recovery transformer block in the pruned model.
For an input sample $\z \sim \mathcal{D}$, let $\h_{\ell}(\z) \in \R^{t \times d}$ denote the hidden state entering the $\ell$-th transformer block, where $t$ is the sequence length and $d$ is the hidden dimension.
For brevity, we write $\h_{\ell}$ for $\h_{\ell}(\z)$ when the dependence on $\z$ is clear.
Suppose blocks $p$ through $p+n-2$ are pruned, while block $p+n-1$ is retained as the recovery module.
Standard layer-wise distillation trains this retained block, denoted by $\sF(\cdot;\theta)$ with pretrained parameters $\theta$, to map the hidden state before the pruned segment directly to the hidden state after the segment:
\begin{equation}
\label{eq:standard}
\sL_{\mathrm{std}}(\theta)=
\mathbb{E}_{\z}
\left[ \left\|\sF(\h_{p};\theta)-\h_{p+n}\right\|_F^2 \right].
\end{equation}
This objective is optimized using a single standard transformer block to approximate the composite transformation originally realized by $n$ consecutive blocks. In practice, such compression often induces a substantial \pterm bottleneck. The parameterization of $\theta$ may not adequately capture the complex nonlinear relationship between $\h_{p}$ and $\h_{p+n}$.

\subsection{Overcomplete Re-parameterization for Scaling Recovery Capacity}
\ours addresses this bottleneck by replacing the compact recovery block during training with an Overcomplete Recovery Module (ORM). The central principle is \emph{``train overcomplete, deploy compact''}: the ORM is temporarily overparameterized during training, but is algebraically folded back into the original compact architecture after training.
Concretely, \ours uses the pretrained parameters $\theta$ with auxiliary parameters $\phi$, yielding an overcomplete re-parameterization $\sFOurs(\cdot;\theta,\phi)$. The recovery objective is defined as:
\begin{equation}
\label{eq:ours}
\sL_{\mathrm{ours}}(\theta,\phi)=
\mathbb{E}_{\z}
\left[ \left\|\sFOurs(\h_{p};\theta,\phi)-\h_{p+n}\right\|_F^2 \right].
\end{equation}
To optimize $\sL_{\mathrm{ours}}$, \ours either freezes $\theta$ and optimizes only $\phi$, or jointly fine-tunes $\theta$ and $\phi$ to increase recovery flexibility.
For notational simplicity, we present \ours with a single recovery block, although $\sFOurs$ may denote a multi-block recovery module. In our experiments, we use a two-block instantiation, as detailed in \cref{sec:practical-implementation}.

By introducing $\phi$, the parameter space is substantially enlarged, facilitating more accurate reconstruction of the target hidden state $\h_{p+n}$. 
Importantly, $\phi$ is structured to be algebraically integrated with $\theta$ into a corresponding compact parameter set $\hat{\theta}$ through re-parameterization $\mathcal{R}$ after training as:
\begin{equation}
\label{eq:ours_deploy}
\hat{\theta} = \mathcal{R}(\theta,\phi), \quad \sFOurs(\h;\theta, \phi) \equiv \sF(\h;\hat{\theta}), \; \forall\, \h.
\end{equation}
Accordingly, \ours increases training-time representational capacity while preserving the original architecture and incurring no additional parameters, memory, or inference-time latency overhead.

\subsection{Overcomplete Recovery Module}
The ORM enlarges the recovery module's optimization space during training without adding inference-time architectural overhead.
Let $\sF(\cdot;\theta)$ denote the standard recovery block, which contains a set of linear projection matrices as:
\begin{gather}
\begin{aligned}
\theta &= \left\{ \P_i : i \in \mathcal{I} \right\}, \\
\mathcal{I} &= \{q,k,v,o,gate,up,down\},
\end{aligned}
\end{gather}
where $q$, $k$, $v$, and $o$ denote the attention projections, while $gate$, $up$, and $down$ denote the feed-forward projections. 
Each projection $\P_i \in \R^{d_i^{\mathrm{out}}\times d_i^{\mathrm{in}}}$ may have its own input and output dimensions depending on the architecture.
To construct the ORM $\sFOurs(\cdot;\theta,\phi)$, \ours expands each base parameter $\P_i$ with an auxiliary parameter pair $(\W_i, \D_i)$. During training, the standard linear projection is replaced by an overcomplete mapping. For an input $\x \in \R^{d_i^\mathrm{in}}$, the overcomplete projection is:
\begin{equation}
\label{eq:overcomplete-projection}
\sFOurs(\x;\;\P_i,\W_i,\D_i) = \D_i(\P_i+\W_i) \x,
\end{equation}
where $\W_i \in \R^{d_i^{\mathrm{out}}\times d_i^{\mathrm{in}}}$ serves as an additive parallel branch that broadens the optimization space, and $\D_i \in \R^{d_i^{\mathrm{out}}\times d_i^{\mathrm{out}}}$ functions as a cascaded multiplicative transformation applied in the output space. Together, these auxiliary parameters improve optimization flexibility during training.
To ensure that the ORM reduces to the original module at initialization, we set $\W_i = \mathbf{0}$ and $\D_i = \mathbf{I}$. This identity initialization guarantees that recovery begins from the pretrained operating point.

\paragraph{Annealed Activation}
Exact algebraic merging requires the ORM to be linear at deployment, but keeping it linear throughout recovery limits training-time flexibility. \ours therefore introduces an \emph{annealed activation} that provides nonlinear training dynamics while gradually returning to the identity map before re-parameterization. At training step $s$, \ours replaces \cref{eq:overcomplete-projection} with
\begin{equation}
\label{eq:overcomplete-projection-with-va}
\sFOursAct{s}(\x;\;\P_i,\W_i,\D_i) = \D_i \AAct_s((\P_i+\W_i) \x),
\end{equation}
where $\AAct_s$ interpolates between the recovery model's activation function $\sigma$ and the identity map:
\begin{equation}
\label{eq:va}
\AAct_s(\x) = \alpha_s \cdot \sigma(\x) + (1-\alpha_s)\x.
\end{equation}
The coefficient $\alpha_s \in [0,1]$ follows a warm-up cosine schedule with a final linear-stabilization phase. Let $S$ be the total number of recovery steps, and let $\rho_{\mathrm{w}}$ and $\rho_{\mathrm{l}}$ denote the warm-up ratio and the final linear-stabilization ratio. We define
\begin{equation}
\alpha_s = \begin{cases} s/S_{\mathrm{w}}, & 0 \le s < S_{\mathrm{w}}, \\
\frac{1}{2}\left(1+\cos\left(\pi\frac{s-S_{\mathrm{w}}}{S_{\mathrm{d}}-S_{\mathrm{w}}}\right)\right), & S_{\mathrm{w}} \le s < S_{\mathrm{d}}, \\
0, & S_{\mathrm{d}} \le s \le S,
\end{cases}
\label{eq:alpha-schedule}
\end{equation}
where $S_{\mathrm{w}}=\rho_{\mathrm{w}}S$ and $S_{\mathrm{d}}=(1-\rho_{\mathrm{l}})S$. Starting from $\alpha_0=0$ ensures that recovery begins from the pretrained operating point, while the warm-up phase rapidly introduces nonlinear training dynamics. The final $\rho_{\mathrm{l}}$ fraction of training keeps $\alpha_s=0$, allowing the weights to stabilize in the exact linear regime before re-parameterization. Thus, at step $S$, $\AAct_S(\x)=\x$, and \cref{eq:overcomplete-projection-with-va} reduces to
\begin{equation}
\sFOursAct{S}(\x;\P_i,\W_i,\D_i)=\D_i(\P_i+\W_i)\x.
\end{equation}
\paragraph{Re-parameterization}
The re-parameterization operation $\sR$ merges each triplet $\{\P_i,\W_i,\D_i\}$ into a single weight matrix $\hat{\P}_i$:
\begin{equation}
\label{eq:reparam}
\hat{\P}_i = \D_i(\P_i + \W_i), \quad \forall i \in \mathcal{I}.
\end{equation}

The pruned model is therefore trained using $\sFOursAct{s}(\h;\theta,\phi)$ but deployed as the original compact module $\sF(\h;\hat{\theta})$. In this way, \ours benefits from enhanced recovery capacity while preserving the inference efficiency of the pruned model.

\paragraph{Optimization geometry of the merged projection}
Although the merged model has the same architecture and function class as a directly fine-tuned one, \ours changes the training parameterization and optimization geometry of the compact deployed weight. For a projection in the final linear phase, let $\M_i=\P_i+\W_i$ and $\hat{\P}_i=\D_i\M_i$, and let $\mathbf{G}$ denote the gradient of $\sL$ with respect to the projection weight in use, i.e., $\partial\sL_{\mathrm{ours}}/\partial\hat{\P}_i$ for \ours and $\partial\sL/\partial\P_i$ for direct fine-tuning. Under continuous-time gradient flow on $\D_i$ and $\W_i$,
\begin{equation}
\label{eq:gradflow}
\frac{d\hat{\P}_i}{dt} = -\,(\D_i\D_i^{\top})\,\mathbf{G} \;-\; \mathbf{G}\,(\M_i^{\top}\M_i),
\end{equation}
whereas directly optimizing $\P_i$ without $\W_i$ and $\D_i$ gives $d\P_i/dt=-\mathbf{G}$. 
The factorized parameterization therefore induces parameter-dependent, positive-semidefinite preconditioning on both sides of the gradient. At identity/zero initialization, where $\D_i=\mathbf{I}$ and $\W_i=\mathbf{0}$, \cref{eq:gradflow} reduces to $-\mathbf{G}-\mathbf{G}\P_i^{\top}\P_i$. Thus, the initial update is shaped by the spectrum of the pretrained projection. Consequently, \ours does not merely add trainable parameters. It optimizes the same compact weight under a different geometry that depends on the pretrained weight, analogous to the mechanism studied in overparameterized deep linear networks~\cite{arora2018overparameterization}. This theoretical result is limited to a frozen-$\P_i$, continuous-gradient-flow model of the final linear phase and does not establish universal superiority or directly model AdamW or nonlinear activation annealing. The annealed activation provides a complementary benefit that the theoretical result does not capture. While $\alpha_s>0$, \ours optimizes over a nonlinear hypothesis class before converging to the linear regime required for exact merging, whose benefit is shown in \cref{sec:ablation}. \Cref{sec:analysis} empirically separates parameterization from capacity using parameter-matched LoRA.

\figBenchmark
\section{Experiments}
We conduct comprehensive experiments on both reasoning and generation benchmarks to compare \ours with state-of-the-art structured pruning methods. 
For pretrained backbones, we use LLaMA2~\cite{touvron2023llama}, LLaMA3~\cite{grattafiori2024llama}, and Qwen3~\cite{qwen3}. We compare against representative recent pruning approaches, including LaCo~\cite{yang2024laco}, ShortGPT~\cite{men2025shortgpt}, and Streamline~\cite{chen2024streamlining}.
Additionally, to directly assess the effectiveness of \ours in the recovery stage, we conduct controlled comparisons under an identical pruning criterion with recent recovery methods, including LoRA~\cite{hu2022lora}, AdaLoRA~\cite{zhang2023adalora}, RankAdaptor~\cite{zhou2025rankadaptor}, and RestoreLCC~\cite{feng2025restoring}.
This isolates the recovery strategy from the pruning criterion, enabling a fair comparison and showing that \ours outperforms existing recovery approaches.

We standardize both the datasets and the evaluation framework across all methods. Specifically, for recovery, we use a fixed subset of the \texttt{FineWeb-Edu}~\cite{lozhkov2024finewebedu} dataset consisting of 120,000 training samples and 4,000 test samples. For evaluation, we adopt the widely used \texttt{lm-eval-harness}~\cite{lmevalharness} framework. This standardized setup enables fairer, more controlled comparisons across methods.
Implementation details are provided in \cref{sec:implementation-details}.

\tabReasoningAdapterFull
\subsection{Training Efficiency and Resource Analysis}
We analyze the relationship among trainable capacity, training resources, throughput, and recovery accuracy in \cref{fig:benchmark}.\footnote{All analyses are conducted with a batch size of 1 and a sequence length of 1024 on two RTX 3090 GPUs (24GB).}
Although \ours introduces substantially more trainable parameters, this increase does not translate into a proportional increase in training cost.
On both backbones, \ours remains comparable to, or only slightly higher than, existing recovery baselines in peak memory and TFLOPs, while achieving higher throughput and stronger reasoning accuracy.
This is because \ours localizes the recovery module after the pruned segment, providing a larger optimization space without requiring backpropagation through long autograd paths, as shown in \cref{fig:main}.
These results indicate that trainable parameter count alone is an incomplete proxy for recovery cost.

\ours also supports a cached variant, \ours{}$^{\ast}$, which reuses precomputed activations from the frozen prefix before the recovery module.
As shown in \cref{fig:benchmark}, this reduces peak GPU memory and TFLOPs by up to $1.6\times$ and $2.4\times$, respectively, while improving throughput by up to $2.8\times$ over vanilla \ours.
Since caching preserves the recovery objective and the final re-parameterized model, \ours{}$^{\ast}$ achieves the same accuracy as \ours while providing additional efficiency gains.

\subsection{Reasoning Performance}
We evaluate reasoning performance under two settings at two pruning ratios (PR). \Cref{tab:reasoning-adapter-full} reports task-level raw accuracy and average retained performance (RP)\footnote{$\text{RP} = \frac{\text{Pruned model average accuracy}}{\text{Dense model average accuracy}} \times 100$} for controlled recovery methods, showing how each recovery strategy behaves across benchmarks.
\Cref{tab:reasoning} shows average RP for complete pruning pipelines, enabling comparison against methods that use their own pruning and recovery procedures.
Detailed results of each task are provided in \cref{sec:detailed-results-reasoning}.

\paragraph{Controlled recovery}
We isolate the effect of recovery by applying all methods to the same pruned model, using a shared pruning criterion that removes the last blocks except the final block.
As shown in \cref{tab:reasoning-adapter-full}, \ours generally achieves the strongest performance across backbones and pruning ratios, with clear gains at aggressive pruning ratios.
In terms of RP, \ours achieves the best result in all controlled settings, improving over the strongest recovery baseline by up to 5.5 points at 25\% pruning and by up to 8.4 points at 50\%.

\tabReasoning
\paragraph{Complete pruning pipeline}
We compare \ours with complete pruning pipelines that use their own pruning and recovery procedures.
As shown in \cref{tab:reasoning}, \ours achieves the strongest RP across all backbones and pruning ratios, suggesting that a fixed pruning mask combined with sufficient recovery capacity can outperform specialized pruning designs. Compatibility with other criteria is examined in \cref{sec:app-generality}.

The improvements are especially pronounced under aggressive pruning, supporting our hypothesis that increased training-time recovery capacity mitigates the \pterm.

\subsection{Generation Performance}
We further evaluate \ours on generation benchmarks to examine whether increased recovery capacity also benefits reasoning-intensive generation. \Cref{tab:generation} reports the average RP over three generation benchmarks~\cite{reddy2019coqa,cobbe2021gsm8k,joshi2017triviaqa}, with detailed raw performance provided in \cref{sec:detailed-results-generation}.

\paragraph{Controlled recovery}
\ours achieves stronger generation performance than controlled recovery baselines across all backbones and pruning ratios. The gains are especially clear under 50\% pruning, where the recovery module must restore more removed knowledge. This suggests that \ours is not limited to multiple-choice reasoning tasks, but also improves recovery on generation benchmarks.

\paragraph{Complete pruning pipeline}
In the generation benchmarks, \ours remains highly competitive against complete pruning pipelines. At 25\% pruning, \ours achieves the best RP across all backbones, and at 50\% pruning, it achieves the best RP on five of six backbones.

Overall, the generation results provide complementary evidence to the reasoning benchmarks. \ours consistently outperforms recovery baselines and remains competitive with complete pruning pipelines, indicating that increased training-time recovery capacity benefits both reasoning and generative evaluation settings.

\tabGeneration

\tabAblation
\subsection{Ablation Study}
\label{sec:ablation}

We conduct ablation studies on LLaMA3-3B with 25\% pruning to analyze the contribution of each \ours component.
\Cref{tab:ablation} compares different ORM configurations by varying the application of $\W_i$, $\D_i$, and the annealed activation $\AAct_s$. We measure training time without feature caching. Full application applies each component to the corresponding projections in both recovery blocks, whereas partial application applies each component only to selected projections in the second recovery block, whose pretrained parameters are frozen.

The results show that MLP-side overcomplete components are more effective than attention-only components, suggesting that feed-forward transformations play a central role in recovering the mappings removed by layer pruning.
We also find that $\W_i$ and $\D_i$ are complementary, as using both components yields stronger performance than using either one alone.
Finally, the hybrid configuration with frozen-block partial application and annealed activation achieves the best overall performance, indicating that selective capacity expansion in the frozen block, combined with nonlinear training dynamics, provides an effective trade-off between recovery flexibility and training cost.

\subsection{Analysis}
\label{sec:analysis}
\tabIsoParameter
\paragraph{Iso-parameter LoRA}
To empirically separate parameterization from capacity, we additionally conduct a LoRA rank-scaling study, keeping the protocol fixed and preserving the standard scaling $\alpha=2r$ at every rank. As shown in \cref{tab:isoparameter}, even under an iso-parameter budget, high-rank LoRA does not close the performance gap with \ours and requires up to 2.9$\times$ more GPU-hours.

\tabFullFTKD
\paragraph{Full capacity controls}
We evaluate full fine-tuning (FT) of the retained blocks and full-model logit-level distillation from the dense teacher. We also combine the same KD objective with \ours, denoted \ours-KD, which replaces the layer-wise MSE objective with logit-level distillation while keeping the \ours parameterization unchanged. As shown in \cref{tab:fullft-and-kd}, full fine-tuning still trails \ours in reasoning at both pruning ratios despite its higher cost, indicating that the recovery parameterization provides benefits beyond exposing more trainable parameters. \ours-KD also outperforms the same logit-KD objective without \ours while using less than one-quarter of the GPU-hours, demonstrating that \ours and logit-level distillation are complementary.

\paragraph{Capacity-knowledge asymmetry in practice}
We further examine how trainable recovery capacity relates to post-pruning performance using the recovery ratio (RR)\footnote{$\mathrm{RR} = \frac{\mathrm{Trainable\ parameters}}{\mathrm{Pruned\ parameters}} \times 100$}.
A small RR indicates limited trainable capacity to compensate for the transformations removed by pruning.

As shown in \cref{fig:analysis}, larger RR values are generally associated with higher retained performance on reasoning and generation benchmarks, with positive rank-based and linear correlations.
This trend is consistent with the capacity-knowledge asymmetry hypothesis, although it should be interpreted as descriptive evidence, as other design factors may also affect performance.
In \ours, RR increases only during recovery through overcomplete parameterization, and we merge the additional parameters before deployment.
Thus, higher training-time recovery capacity adds no inference-time parameters, memory, or latency overhead.

\figAnalysis
\section{Conclusion}
We presented Overcomplete Reparameterization (\ours), a post-pruning recovery framework for structured LLM pruning.
Motivated by capacity-knowledge asymmetry, we revisit the recovery stage as a capacity-limited reconstruction problem, where a compact trainable module must recover the transformations pruned away.
\ours addresses this bottleneck under the principle of \emph{``train overcomplete, deploy compact''}: temporarily increasing training-time recovery capacity through an overcomplete recovery module, while preserving the compact inference-time architecture via algebraic re-parameterization.
Its auxiliary components enlarge the optimization space, and the annealed activation enables nonlinear training dynamics before converging to a linear regime that permits merging.

Across multiple backbones, \ours improves performance without increasing inference-time overhead. Our analysis shows that increased recovery capacity does not necessarily require a proportional increase in training cost.
These results suggest training-time overcomplete parameterization as an effective and deployment-friendly strategy for structured LLM pruning.

\section*{Limitations}
\ours focuses on improving the recovery stage of structured pruning, with main results based on layer pruning. This setting provides a controlled testbed for studying recovery capacity. However, the channel-wise pruning and hybrid-architecture results in \cref{sec:app-generality} remain preliminary because they use smaller recovery budgets than the main protocol and cover only one or two backbones each. Attention-head and mixed structured pruning remain unexplored. A comprehensive evaluation across pruning granularity is left for future work.

\ours also increases the number of trainable parameters during recovery.
Although our efficiency analysis shows that this expansion does not proportionally increase the practical training cost and can be further accelerated through feature caching, the cached variant assumes a frozen, reusable prefix before the recovery module.
Thus, its benefit may vary with pruning patterns, hardware, and training implementations.
Future work could extend overcomplete recovery to more diverse pruning structures and hardware settings.

\section*{Acknowledgments}
This research was supported by the National Research Foundation of Korea (NRF), Electronics and Telecommunications Research Institute (ETRI), and Institute of Information \& Communications
Technology Planning \& Evaluation (IITP), funded by the Ministry of Education (RS-2025-25423987), the Korean government [26CS1100, Development of Proprietary Physical AI-based Small-scale Computers and Integrated Soft Suits], and the Korean government (MSIT) (RS-2026-25518808, RS-2026-25617480, and IITP-2026-RS-2023-00255968).

\bibliography{ref}

\begin{thebibliography}{39}
\providecommand{\natexlab}[1]{#1}

\bibitem[{Amini et~al.(2019)Amini, Gabriel, Lin, Koncel-Kedziorski, Choi, and Hajishirzi}]{amini2019mathqa}
Aida Amini, Saadia Gabriel, Shanchuan Lin, Rik Koncel-Kedziorski, Yejin Choi, and Hannaneh Hajishirzi. 2019.
\newblock \href {https://doi.org/10.18653/v1/N19-1245} {{M}ath{QA}: Towards interpretable math word problem solving with operation-based formalisms}.
\newblock In \emph{Proceedings of the 2019 Conference of the North American Chapter of the Association for Computational Linguistics: Human Language Technologies, Volume 1 (Long and Short Papers)}, pages 2357--2367.

\bibitem[{An et~al.(2024)An, Zhao, Yu, Tang, and Wang}]{an2024flap}
Yongqi An, Xu~Zhao, Tao Yu, Ming Tang, and Jinqiao Wang. 2024.
\newblock \href {https://doi.org/10.1609/aaai.v38i10.28960} {Fluctuation-based adaptive structured pruning for large language models}.
\newblock In \emph{Proceedings of the AAAI Conference on Artificial Intelligence}, volume~38, pages 10865--10873. AAAI Press.

\bibitem[{Arora et~al.(2018)Arora, Cohen, and Hazan}]{arora2018overparameterization}
Sanjeev Arora, Nadav Cohen, and Elad Hazan. 2018.
\newblock \href {https://proceedings.mlr.press/v80/arora18a.html} {On the optimization of deep networks: Implicit acceleration by overparameterization}.
\newblock In \emph{Proceedings of the 35th International Conference on Machine Learning}, volume~80 of \emph{Proceedings of Machine Learning Research}, pages 244--253. PMLR.

\bibitem[{Ashkboos et~al.(2024)Ashkboos, Croci, Nascimento, Hoefler, and Hensman}]{ashkboos2024slicegpt}
Saleh Ashkboos, Maximilian~L Croci, Marcelo Gennari~do Nascimento, Torsten Hoefler, and James Hensman. 2024.
\newblock \href {https://openreview.net/forum?id=vXxardq6db} {{S}lice{GPT}: Compress large language models by deleting rows and columns}.
\newblock In \emph{International Conference on Learning Representations}.

\bibitem[{Bisk et~al.(2020)Bisk, Zellers, Bras, Gao, and Choi}]{bisk2020piqa}
Yonatan Bisk, Rowan Zellers, Ronan~Le Bras, Jianfeng Gao, and Yejin Choi. 2020.
\newblock \href {https://doi.org/10.1609/aaai.v34i05.6239} {{PIQA}: Reasoning about physical commonsense in natural language}.
\newblock In \emph{Proceedings of the AAAI Conference on Artificial Intelligence}, volume~34, pages 7432--7439. AAAI Press.

\bibitem[{Blakeman et~al.(2025)Blakeman, Basant, Khattar, Renduchintala, Bercovich, Ficek, Bjorlin, Taghibakhshi, Deshmukh, Mahabaleshwarkar, Tao, Shors, Aithal, Poojary, Dattagupta, Buddharaju, Chen, Ginsburg, Wang, Norick, Butterfield, Catanzaro, del Mundo, Dong, Harvey, Parisien, Su, Korzekwa, Yin, Gitman, Mosallanezhad, Narayanan, Fridman, Rekesh, Ma, Pykhtar, Ahn, Riach, Stosic, Long, Segal, Evans, Chung, Galinkin, Bakhturina, Dobrowolska, Jia, Liu, Prasad, Shen, Liu, Chen, Qian, Ngo, Liu, Li, Gitman, Karmanov, Moshkov, Golan, Kautz, Scowcroft, Casper, Seppanen, Lu, Sewall, Zeng, You, Zhang, Zhang, Huang, Xue, Huang, Conway, Kamalu, Barker, Cohen, Jennings, Parmar, Sapra, Briski, Chumachenko, Luna, Santhanam, Kong, Sivamani, Pawelec, Anik, Li, McAfee, Derczynski, Pavao, Vega, Voegtle, Bala, de~Melo, Sreedhar, Chochowski, Kliegl, Stepniewska-Dziubinska, Le, Novikov, Samadi, Andersch, Evans, Martinez, Chrzanowski, Ranzinger, Blaz, Smelyanskiy, Fawzy, Shoeybi, Patwary, Lee, Tajbakhsh, Xu, Rybakov, Kuchaiev,
  Delalleau, Nitski, Chadha, Shamis, Micikevicius, Molchanov, Dykas, Fischer, Aquilanti, Bialecki, Varshney, Gundecha, Tredak, Karimi, Kandu, El-Yaniv, Joshi, Waleffe, Zhang, Kavanaugh, Jain, Kriman, Lym, Satheesh, Muralidharan, Narenthiran, Anandaraj, Bak, Kashirsky, Han, Acharya, Ghosh, Sreenivas, Clay, Thomas, Prabhumoye, Pachori, Toshniwal, Prayaga, Jain, Das, Kierat, Majumdar, Han, Singhal, Niverty, Alborghetti, Panguluri, Bhendigeri, Akter, Migacz, Shiri, Kong, Roman, Ronen, Saar, Konuk, Rintamaki, Poon, De, Noroozi, Singh, Korthikanti, Kurin, Ahmad, Du, Ping, Dai, Byeon, Ren, Xu, Choi, Zhang, Lin, Suhara, Yu, Li, Li, Zhu, Yang, and Chen}]{taghibakhshi2025nemotron}
Aaron Blakeman, Aarti Basant, Abhinav Khattar, Adithya Renduchintala, Akhiad Bercovich, Aleksander Ficek, Alexis Bjorlin, Ali Taghibakhshi, Amala~Sanjay Deshmukh, Ameya~Sunil Mahabaleshwarkar, Andrew Tao, Anna Shors, Ashwath Aithal, Ashwin Poojary, Ayush Dattagupta, Balaram Buddharaju, Bobby Chen, Boris Ginsburg, Boxin Wang, and 180 others. 2025.
\newblock \href {https://arxiv.org/abs/2504.03624} {Nemotron-h: A family of accurate and efficient hybrid mamba-transformer models}.
\newblock \emph{Preprint}, arXiv:2504.03624.

\bibitem[{Chen et~al.(2025)Chen, Hu, Zhang, Wang, Li, and Chen}]{chen2024streamlining}
Xiaodong Chen, Yuxuan Hu, Jing Zhang, Yanling Wang, Cuiping Li, and Hong Chen. 2025.
\newblock \href {https://openreview.net/forum?id=IC5RJvRoMp} {Streamlining redundant layers to compress large language models}.
\newblock In \emph{International Conference on Learning Representations}.

\bibitem[{Clark et~al.(2019)Clark, Lee, Chang, Kwiatkowski, Collins, and Toutanova}]{clark2019boolq}
Christopher Clark, Kenton Lee, Ming-Wei Chang, Tom Kwiatkowski, Michael Collins, and Kristina Toutanova. 2019.
\newblock \href {https://doi.org/10.18653/v1/N19-1300} {{B}ool{Q}: Exploring the surprising difficulty of natural yes/no questions}.
\newblock In \emph{Proceedings of the 2019 Conference of the North American Chapter of the Association for Computational Linguistics: Human Language Technologies, Volume 1 (Long and Short Papers)}, pages 2924--2936.

\bibitem[{Clark et~al.(2018)Clark, Cowhey, Etzioni, Khot, Sabharwal, Schoenick, and Tafjord}]{peter2018arc}
Peter Clark, Isaac Cowhey, Oren Etzioni, Tushar Khot, Ashish Sabharwal, Carissa Schoenick, and Oyvind Tafjord. 2018.
\newblock \href {https://doi.org/10.48550/arXiv.1803.05457} {Think you have solved question answering? try {ARC}, the {AI}2 reasoning challenge}.
\newblock \emph{arXiv preprint arXiv:1803.05457}.

\bibitem[{Cobbe et~al.(2021)Cobbe, Kosaraju, Bavarian, Chen, Jun, Kaiser, Plappert, Tworek, Hilton, and Nakano}]{cobbe2021gsm8k}
Karl Cobbe, Vineet Kosaraju, Mohammad Bavarian, Mark Chen, Heewoo Jun, Lukasz Kaiser, Matthias Plappert, Jerry Tworek, Jacob Hilton, and Reiichiro Nakano. 2021.
\newblock \href {https://doi.org/10.48550/arXiv.2110.14168} {Training verifiers to solve math word problems}.
\newblock \emph{arXiv preprint arXiv:2110.14168}.

\bibitem[{Dao and Gu(2024)}]{dao2024transformers}
Tri Dao and Albert Gu. 2024.
\newblock \href {https://arxiv.org/abs/2405.21060} {Transformers are ssms: Generalized models and efficient algorithms through structured state space duality}.
\newblock \emph{arXiv preprint arXiv:2405.21060}.

\bibitem[{Ding et~al.(2019)Ding, Guo, Ding, and Han}]{acnet}
Xiaohan Ding, Yuchen Guo, Guiguang Ding, and Jungong Han. 2019.
\newblock \href {https://doi.org/10.1109/ICCV.2019.00200} {{ACN}et: Strengthening the kernel skeletons for powerful {CNN} via asymmetric convolution blocks}.
\newblock In \emph{Proceedings of the IEEE/CVF International Conference on Computer Vision}, pages 1911--1920.

\bibitem[{Ding et~al.(2021{\natexlab{a}})Ding, Zhang, Han, and Ding}]{dbb}
Xiaohan Ding, Xiangyu Zhang, Jungong Han, and Guiguang Ding. 2021{\natexlab{a}}.
\newblock \href {https://doi.org/10.1109/CVPR46437.2021.01074} {Diverse branch block: Building a convolution as an inception-like unit}.
\newblock In \emph{Proceedings of the IEEE/CVF Conference on Computer Vision and Pattern Recognition}, pages 10886--10895.

\bibitem[{Ding et~al.(2021{\natexlab{b}})Ding, Zhang, Ma, Han, Ding, and Sun}]{repvgg}
Xiaohan Ding, Xiangyu Zhang, Ningning Ma, Jungong Han, Guiguang Ding, and Jian Sun. 2021{\natexlab{b}}.
\newblock \href {https://arxiv.org/abs/2101.03697} {{R}ep{VGG}: Making {VGG}-style {C}onv{N}ets great again}.
\newblock In \emph{Proceedings of the IEEE/CVF Conference on Computer Vision and Pattern Recognition}, pages 13728--13737.

\bibitem[{Feng et~al.(2025)Feng, Zhou, Zhu, Li, Chua, Mak, Ng, and Mao}]{feng2025restoring}
Zijian Feng, Hanzhang Zhou, Zixiao Zhu, Tianjiao Li, Jia Jim~Deryl Chua, Lee~Onn Mak, Gee~Wah Ng, and Kezhi Mao. 2025.
\newblock \href {https://openreview.net/forum?id=cECo8tetzF} {Restoring pruned large language models via lost component compensation}.
\newblock In \emph{Advances in Neural Information Processing Systems}.

\bibitem[{Gao et~al.(2024)Gao, Tow, Abbasi, Biderman, Black, DiPofi, Foster, Golding, Hsu, Le~Noac'h, Li, McDonell, Muennighoff, Ociepa, Phang, Reynolds, Schoelkopf, Skowron, Sutawika, Tang, Thite, Wang, Wang, and Zou}]{lmevalharness}
Leo Gao, Jonathan Tow, Baber Abbasi, Stella Biderman, Sid Black, Anthony DiPofi, Charles Foster, Laurence Golding, Jeffrey Hsu, Alain Le~Noac'h, Haonan Li, Kyle McDonell, Niklas Muennighoff, Chris Ociepa, Jason Phang, Laria Reynolds, Hailey Schoelkopf, Aviya Skowron, Lintang Sutawika, and 5 others. 2024.
\newblock \href {https://doi.org/10.5281/zenodo.12608602} {The language model evaluation harness}.

\bibitem[{Grattafiori et~al.(2024)Grattafiori, Dubey, Jauhri, Pandey, Kadian, Al-Dahle, Letman, Mathur, Schelten, Vaughan, Yang, Fan, Goyal, Hartshorn, Yang, Mitra, Sravankumar, Korenev, Hinsvark, Rao, Zhang, Rodriguez, Gregerson, Spataru, Roziere, Biron, Tang, Chern, Caucheteux, Nayak, Bi, Marra, McConnell, Keller, Touret, Wu, Wong, Ferrer, Nikolaidis, Allonsius, Song, Pintz, Livshits, Wyatt, Esiobu, Choudhary, Mahajan, Garcia-Olano, Perino, Hupkes, Lakomkin, AlBadawy, Lobanova, Dinan, Smith, Radenovic, Guzmán, Zhang, Synnaeve, Lee, Anderson, Thattai, Nail, Mialon, Pang, Cucurell, Nguyen, Korevaar, Xu, Touvron, Zarov, Ibarra, Kloumann, Misra, Evtimov, Zhang, Copet, Lee, Geffert, Vranes, Park, Mahadeokar, Shah, van~der Linde, Billock, Hong, Lee, Fu, Chi, Huang, Liu, Wang, Yu, Bitton, Spisak, Park, Rocca, Johnstun, Saxe, Jia, Alwala, Prasad, Upasani, Plawiak, Li, Heafield, Stone, El-Arini, Iyer, Malik, Chiu, Bhalla, Lakhotia, Rantala-Yeary, van~der Maaten, Chen, Tan, Jenkins, Martin, Madaan, Malo, Blecher,
  Landzaat, de~Oliveira, Muzzi, Pasupuleti, Singh, Paluri, Kardas, Tsimpoukelli, Oldham, Rita, Pavlova, Kambadur, Lewis, Si, Singh, Hassan, Goyal, Torabi, Bashlykov, Bogoychev, Chatterji, Zhang, Duchenne, Çelebi, Alrassy, Zhang, Li, Vasic, Weng, Bhargava, Dubal, Krishnan, Koura, Xu, He, Dong, Srinivasan, Ganapathy, Calderer, Cabral, Stojnic, Raileanu, Maheswari, Girdhar, Patel, Sauvestre, Polidoro, Sumbaly, Taylor, Silva, Hou, Wang, Hosseini, Chennabasappa, Singh, Bell, Kim, Edunov, Nie, Narang, Raparthy, Shen, Wan, Bhosale, Zhang, Vandenhende, Batra, Whitman, Sootla, Collot, Gururangan, Borodinsky, Herman, Fowler, Sheasha, Georgiou, Scialom, Speckbacher, Mihaylov, Xiao, Karn, Goswami, Gupta, Ramanathan, Kerkez, Gonguet, Do, Vogeti, Albiero, Petrovic, Chu, Xiong, Fu, Meers, Martinet, Wang, Wang, Tan, Xia, Xie, Jia, Wang, Goldschlag, Gaur, Babaei, Wen, Song, Zhang, Li, Mao, Coudert, Yan, Chen, Papakipos, Singh, Srivastava, Jain, Kelsey, Shajnfeld, Gangidi, Victoria, Goldstand, Menon, Sharma, Boesenberg,
  Baevski, Feinstein, Kallet, Sangani, Teo, Yunus, Lupu, Alvarado, Caples, Gu, Ho, Poulton, Ryan, Ramchandani, Dong, Franco, Goyal, Saraf, Chowdhury, Gabriel, Bharambe, Eisenman, Yazdan, James, Maurer, Leonhardi, Huang, Loyd, Paola, Paranjape, Liu, Wu, Ni, Hancock, Wasti, Spence, Stojkovic, Gamido, Montalvo, Parker, Burton, Mejia, Liu, Wang, Kim, Zhou, Hu, Chu, Cai, Tindal, Feichtenhofer, Gao, Civin, Beaty, Kreymer, Li, Adkins, Xu, Testuggine, David, Parikh, Liskovich, Foss, Wang, Le, Holland, Dowling, Jamil, Montgomery, Presani, Hahn, Wood, Le, Brinkman, Arcaute, Dunbar, Smothers, Sun, Kreuk, Tian, Kokkinos, Ozgenel, Caggioni, Kanayet, Seide, Florez, Schwarz, Badeer, Swee, Halpern, Herman, Sizov, Guangyi, Zhang, Lakshminarayanan, Inan, Shojanazeri, Zou, Wang, Zha, Habeeb, Rudolph, Suk, Aspegren, Goldman, Zhan, Damlaj, Molybog, Tufanov, Leontiadis, Veliche, Gat, Weissman, Geboski, Kohli, Lam, Asher, Gaya, Marcus, Tang, Chan, Zhen, Reizenstein, Teboul, Zhong, Jin, Yang, Cummings, Carvill, Shepard, McPhie,
  Torres, Ginsburg, Wang, Wu, U, Saxena, Khandelwal, Zand, Matosich, Veeraraghavan, Michelena, Li, Jagadeesh, Huang, Chawla, Huang, Chen, Garg, A, Silva, Bell, Zhang, Guo, Yu, Moshkovich, Wehrstedt, Khabsa, Avalani, Bhatt, Mankus, Hasson, Lennie, Reso, Groshev, Naumov, Lathi, Keneally, Liu, Seltzer, Valko, Restrepo, Patel, Vyatskov, Samvelyan, Clark, Macey, Wang, Hermoso, Metanat, Rastegari, Bansal, Santhanam, Parks, White, Bawa, Singhal, Egebo, Usunier, Mehta, Laptev, Dong, Cheng, Chernoguz, Hart, Salpekar, Kalinli, Kent, Parekh, Saab, Balaji, Rittner, Bontrager, Roux, Dollar, Zvyagina, Ratanchandani, Yuvraj, Liang, Alao, Rodriguez, Ayub, Murthy, Nayani, Mitra, Parthasarathy, Li, Hogan, Battey, Wang, Howes, Rinott, Mehta, Siby, Bondu, Datta, Chugh, Hunt, Dhillon, Sidorov, Pan, Mahajan, Verma, Yamamoto, Ramaswamy, Lindsay, Lindsay, Feng, Lin, Zha, Patil, Shankar, Zhang, Zhang, Wang, Agarwal, Sajuyigbe, Chintala, Max, Chen, Kehoe, Satterfield, Govindaprasad, Gupta, Deng, Cho, Virk, Subramanian, Choudhury,
  Goldman, Remez, Glaser, Best, Koehler, Robinson, Li, Zhang, Matthews, Chou, Shaked, Vontimitta, Ajayi, Montanez, Mohan, Kumar, Mangla, Ionescu, Poenaru, Mihailescu, Ivanov, Li, Wang, Jiang, Bouaziz, Constable, Tang, Wu, Wang, Wu, Gao, Kleinman, Chen, Hu, Jia, Qi, Li, Zhang, Zhang, Adi, Nam, Yu, Wang, Zhao, Hao, Qian, Li, He, Rait, DeVito, Rosnbrick, Wen, Yang, Zhao, and Ma}]{grattafiori2024llama}
Aaron Grattafiori, Abhimanyu Dubey, Abhinav Jauhri, Abhinav Pandey, Abhishek Kadian, Ahmad Al-Dahle, Aiesha Letman, Akhil Mathur, Alan Schelten, Alex Vaughan, Amy Yang, Angela Fan, Anirudh Goyal, Anthony Hartshorn, Aobo Yang, Archi Mitra, Archie Sravankumar, Artem Korenev, Arthur Hinsvark, and 542 others. 2024.
\newblock \href {https://doi.org/10.48550/arXiv.2407.21783} {The llama 3 herd of models}.
\newblock \emph{Preprint}, arXiv:2407.21783.

\bibitem[{Hendrycks et~al.(2021)Hendrycks, Burns, Basart, Zou, Mazeika, Song, and Steinhardt}]{hendrycks2020mmlu}
Dan Hendrycks, Collin Burns, Steven Basart, Andy Zou, Mantas Mazeika, Dawn Song, and Jacob Steinhardt. 2021.
\newblock \href {https://arxiv.org/abs/2009.03300} {Measuring massive multitask language understanding}.
\newblock In \emph{International Conference on Learning Representations}.

\bibitem[{Hu et~al.(2022)Hu, Shen, Wallis, Allen-Zhu, Li, Wang, Wang, and Chen}]{hu2022lora}
Edward~J Hu, Yelong Shen, Phillip Wallis, Zeyuan Allen-Zhu, Yuanzhi Li, Shean Wang, Lu~Wang, and Weizhu Chen. 2022.
\newblock \href {https://openreview.net/forum?id=nZeVKeeFYf9} {{L}o{RA}: Low-rank adaptation of large language models}.
\newblock In \emph{International Conference on Learning Representations}.

\bibitem[{Joshi et~al.(2017)Joshi, Choi, Weld, and Zettlemoyer}]{joshi2017triviaqa}
Mandar Joshi, Eunsol Choi, Daniel~S Weld, and Luke Zettlemoyer. 2017.
\newblock \href {https://doi.org/10.18653/v1/P17-1147} {{T}rivia{QA}: A large scale distantly supervised challenge dataset for reading comprehension}.
\newblock In \emph{Proceedings of the 55th Annual Meeting of the Association for Computational Linguistics (Volume 1: Long Papers)}, pages 1601--1611.

\bibitem[{Kopiczko et~al.(2024)Kopiczko, Blankevoort, and Asano}]{kopiczko2023vera}
Dawid~J Kopiczko, Tijmen Blankevoort, and Yuki~M Asano. 2024.
\newblock \href {https://openreview.net/forum?id=NjNfLdxr3A} {Ve{RA}: Vector-based random matrix adaptation}.
\newblock In \emph{International Conference on Learning Representations}.

\bibitem[{Lai et~al.(2017)Lai, Xie, Liu, Yang, and Hovy}]{lai2017race}
Guokun Lai, Qizhe Xie, Hanxiao Liu, Yiming Yang, and Eduard Hovy. 2017.
\newblock \href {https://doi.org/10.18653/v1/D17-1082} {{RACE}: Large-scale {R}e{A}ding comprehension dataset from examinations}.
\newblock In \emph{Proceedings of the 2017 Conference on Empirical Methods in Natural Language Processing}, pages 785--794.

\bibitem[{Lozhkov et~al.(2024)Lozhkov, Ben~Allal, von Werra, and Wolf}]{lozhkov2024finewebedu}
Anton Lozhkov, Loubna Ben~Allal, Leandro von Werra, and Thomas Wolf. 2024.
\newblock \href {https://huggingface.co/datasets/HuggingFaceFW/fineweb-edu} {Fineweb-edu: the finest collection of educational content}.

\bibitem[{Ma et~al.(2023)Ma, Fang, and Wang}]{ma2023llmpruner}
Xinyin Ma, Gongfan Fang, and Xinchao Wang. 2023.
\newblock \href {https://proceedings.neurips.cc/paper_files/paper/2023/hash/44956951349095f74492a5471128a7e0-Abstract-Conference.html} {{LLM}-pruner: On the structural pruning of large language models}.
\newblock In \emph{Advances in Neural Information Processing Systems}.

\bibitem[{Men et~al.(2025)Men, Xu, Zhang, Yuan, Wang, Lin, Lu, Han, and Chen}]{men2025shortgpt}
Xin Men, Mingyu Xu, Qingyu Zhang, Qianhao Yuan, Bingning Wang, Hongyu Lin, Yaojie Lu, Xianpei Han, and Weipeng Chen. 2025.
\newblock \href {https://doi.org/10.18653/v1/2025.findings-acl.1035} {{S}hort{GPT}: Layers in large language models are more redundant than you expect}.
\newblock In \emph{Findings of the Association for Computational Linguistics: ACL 2025}, pages 20192--20204, Vienna, Austria. Association for Computational Linguistics.

\bibitem[{Mihaylov et~al.(2018)Mihaylov, Clark, Khot, and Sabharwal}]{mihaylov2018openbookqa}
Todor Mihaylov, Peter Clark, Tushar Khot, and Ashish Sabharwal. 2018.
\newblock \href {https://doi.org/10.18653/v1/D18-1260} {Can a suit of armor conduct electricity? a new dataset for open book question answering}.
\newblock In \emph{Proceedings of the 2018 Conference on Empirical Methods in Natural Language Processing}, pages 2381--2391.

\bibitem[{Oh and Ryu(2024)}]{oh2024neural}
Seungmin Oh and Jongbin Ryu. 2024.
\newblock \href {https://doi.org/10.1007/978-981-96-0966-6_7} {Neural substitution for branch-level network re-parameterization}.
\newblock In \emph{Proceedings of the Asian Conference on Computer Vision}, pages 104--120. Springer.

\bibitem[{Reddy et~al.(2019)Reddy, Chen, and Manning}]{reddy2019coqa}
Siva Reddy, Danqi Chen, and Christopher~D Manning. 2019.
\newblock \href {https://doi.org/10.1162/tacl_a_00266} {{C}o{QA}: A conversational question answering challenge}.
\newblock \emph{Transactions of the Association for Computational Linguistics}, 7:249--266.

\bibitem[{Sakaguchi et~al.(2021)Sakaguchi, Bras, Bhagavatula, and Choi}]{sakaguchi2021winogrande}
Keisuke Sakaguchi, Ronan~Le Bras, Chandra Bhagavatula, and Yejin Choi. 2021.
\newblock \href {https://doi.org/10.1145/3474381} {Winogrande: An adversarial winograd schema challenge at scale}.
\newblock \emph{Communications of the ACM}, 64(9):99--106.

\bibitem[{Touvron et~al.(2023)Touvron, Martin, Stone, Albert, Almahairi, Babaei, Bashlykov, Batra, Bhargava, Bhosale, Bikel, Blecher, Ferrer, Chen, Cucurull, Esiobu, Fernandes, Fu, Fu, Fuller, Gao, Goswami, Goyal, Hartshorn, Hosseini, Hou, Inan, Kardas, Kerkez, Khabsa, Kloumann, Korenev, Koura, Lachaux, Lavril, Lee, Liskovich, Lu, Mao, Martinet, Mihaylov, Mishra, Molybog, Nie, Poulton, Reizenstein, Rungta, Saladi, Schelten, Silva, Smith, Subramanian, Tan, Tang, Taylor, Williams, Kuan, Xu, Yan, Zarov, Zhang, Fan, Kambadur, Narang, Rodriguez, Stojnic, Edunov, and Scialom}]{touvron2023llama}
Hugo Touvron, Louis Martin, Kevin Stone, Peter Albert, Amjad Almahairi, Yasmine Babaei, Nikolay Bashlykov, Soumya Batra, Prajjwal Bhargava, Shruti Bhosale, Dan Bikel, Lukas Blecher, Cristian~Canton Ferrer, Moya Chen, Guillem Cucurull, David Esiobu, Jude Fernandes, Jeremy Fu, Wenyin Fu, and 49 others. 2023.
\newblock \href {https://doi.org/10.48550/arXiv.2307.09288} {Llama 2: Open foundation and fine-tuned chat models}.
\newblock \emph{Preprint}, arXiv:2307.09288.

\bibitem[{Vasu et~al.(2023)Vasu, Gabriel, Zhu, Tuzel, and Ranjan}]{vasu2023fastvit}
Pavan Kumar~Anasosalu Vasu, James Gabriel, Jeff Zhu, Oncel Tuzel, and Anurag Ranjan. 2023.
\newblock \href {https://arxiv.org/abs/2303.14189} {{F}ast{V}i{T}: A fast hybrid vision transformer using structural reparameterization}.
\newblock In \emph{Proceedings of the IEEE/CVF International Conference on Computer Vision}.

\bibitem[{Wang et~al.(2025{\natexlab{a}})Wang, Shen, Ding, Xue, Liu, and Ding}]{wang2025come}
Fei Wang, Li~Shen, Liang Ding, Chao Xue, Ye~Liu, and Changxing Ding. 2025{\natexlab{a}}.
\newblock \href {https://openreview.net/forum?id=enhFXzKii4} {Layer as puzzle pieces: Compressing large language models through layer concatenation}.
\newblock In \emph{Advances in Neural Information Processing Systems}.

\bibitem[{Wang et~al.(2025{\natexlab{b}})Wang, Ma, Wang, Chen, Liping, Yang, Xu, Liu, and Qin}]{wang2025cfsp}
Yuxin Wang, MingHua Ma, Zekun Wang, Jingchang Chen, Shan Liping, Qing Yang, Dongliang Xu, Ming Liu, and Bing Qin. 2025{\natexlab{b}}.
\newblock \href {https://aclanthology.org/2025.coling-main.626/} {{CFSP}: An efficient structured pruning framework for {LLM}s with coarse-to-fine activation information}.
\newblock In \emph{Proceedings of the 31st International Conference on Computational Linguistics}, pages 9311--9328, Abu Dhabi, UAE. Association for Computational Linguistics.

\bibitem[{Yang et~al.(2025)Yang, Li, Yang, Zhang, Hui, Zheng, Yu, Gao, Huang, Lv, Zheng, Liu, Zhou, Huang, Hu, Ge, Wei, Lin, Tang, Yang, Tu, Zhang, Yang, Yang, Zhou, Zhou, Lin, Dang, Bao, Yang, Yu, Deng, Li, Xue, Li, Zhang, Wang, Zhu, Men, Gao, Liu, Luo, Li, Tang, Yin, Ren, Wang, Zhang, Ren, Fan, Su, Zhang, Zhang, Wan, Liu, Wang, Cui, Zhang, Zhou, and Qiu}]{qwen3}
An~Yang, Anfeng Li, Baosong Yang, Beichen Zhang, Binyuan Hui, Bo~Zheng, Bowen Yu, Chang Gao, Chengen Huang, Chenxu Lv, Chujie Zheng, Dayiheng Liu, Fan Zhou, Fei Huang, Feng Hu, Hao Ge, Haoran Wei, Huan Lin, Jialong Tang, and 41 others. 2025.
\newblock \href {https://doi.org/10.48550/arXiv.2505.09388} {{Q}wen3 technical report}.
\newblock \emph{arXiv preprint arXiv:2505.09388}.

\bibitem[{Yang et~al.(2024)Yang, Cao, and Zhao}]{yang2024laco}
Yifei Yang, Zouying Cao, and Hai Zhao. 2024.
\newblock \href {https://doi.org/10.18653/v1/2024.findings-emnlp.372} {{L}a{C}o: Large language model pruning via layer collapse}.
\newblock In \emph{Findings of the Association for Computational Linguistics: EMNLP 2024}, pages 6401--6417.

\bibitem[{Zellers et~al.(2019)Zellers, Holtzman, Bisk, Farhadi, and Choi}]{zellers2019hellaswag}
Rowan Zellers, Ari Holtzman, Yonatan Bisk, Ali Farhadi, and Yejin Choi. 2019.
\newblock \href {https://doi.org/10.18653/v1/P19-1472} {{H}ella{S}wag: Can a machine really finish your sentence?}
\newblock In \emph{Proceedings of the 57th Annual Meeting of the Association for Computational Linguistics}, pages 4791--4800.

\bibitem[{Zhang et~al.(2023)Zhang, Chen, Bukharin, He, Cheng, Chen, and Zhao}]{zhang2023adalora}
Qingru Zhang, Minshuo Chen, Alexander Bukharin, Pengcheng He, Yu~Cheng, Weizhu Chen, and Tuo Zhao. 2023.
\newblock \href {https://openreview.net/forum?id=lq62uWRJjiY} {Adaptive budget allocation for parameter-efficient fine-tuning}.
\newblock In \emph{International Conference on Learning Representations}.

\bibitem[{Zhang et~al.(2024)Zhang, Li, Wang, Shen, Plank, Bischl, Rezaei, and Kawaguchi}]{zhang2024finercut}
Yang Zhang, Yawei Li, Xinpeng Wang, Qianli Shen, Barbara Plank, Bernd Bischl, Mina Rezaei, and Kenji Kawaguchi. 2024.
\newblock \href {https://openreview.net/forum?id=jrSWzgno4W} {Finercut: Finer-grained interpretable layer pruning for large language models}.
\newblock In \emph{Machine Learning and Compression Workshop at NeurIPS 2024}.

\bibitem[{Zhou et~al.(2025)Zhou, Han, Yang, Zhou, Cheng, Wang, and Li}]{zhou2025rankadaptor}
Changhai Zhou, Shijie Han, Lining Yang, Yuhua Zhou, Xu~Cheng, Yibin Wang, and Hongguang Li. 2025.
\newblock \href {https://doi.org/10.18653/v1/2025.findings-naacl.321} {Rankadaptor: Hierarchical rank allocation for efficient fine-tuning pruned {LLM}s via performance model}.
\newblock In \emph{Findings of the Association for Computational Linguistics: NAACL 2025}, pages 5796--5810, Albuquerque, New Mexico. Association for Computational Linguistics.

\end{thebibliography}

\appendix
\crefalias{section}{appendix}
\crefalias{subsection}{appendix}

\section*{Appendix}
Unless otherwise noted, `Rea.' and `Gen.' in the appendix denote the average raw score over the 10 reasoning and 3 generation tasks, not RP.

\section{Practical Implementation}
\label{sec:practical-implementation}
For notational simplicity, the method sections describe \ours using a single recovery block. In our implementation, we instantiate $\sFOurs$ with two consecutive transformer blocks following the pruned segment. This provides additional recovery flexibility while keeping the active training path short. Let $\F_{\mathrm{ORM},1}$ and $\F_{\mathrm{ORM},2}$ denote the first and second \ours recovery blocks, respectively. The practical recovery objective is written as
\begin{equation}
\tilde{\h} =
\F_{\mathrm{ORM},1}(\h_p;\theta_1,\phi_1),
\end{equation}
\begin{equation}
\mathcal{L}_{\mathrm{ours}} = \mathbb{E}_{\z} \left[ \left\| \F_{\mathrm{ORM},2}(\tilde{\h};\bar{\theta}_2,\phi_2) - \h_{p+n+1} \right\|_F^2 \right],
\end{equation}
where $\theta_1$ denotes the pretrained parameters of the first recovery block, $\bar{\theta}_2$ denotes the frozen pretrained parameters of the second recovery block, and $\phi_1,\phi_2$ are the corresponding \ours auxiliary parameters. During recovery, we jointly optimize $\theta_1$ and $\phi_1$ in the first block, keep $\bar{\theta}_2$ frozen, and optimize only $\phi_2$ in the second block. After training, each overcomplete projection in both blocks is independently merged using the same projection-wise re-parameterization rule in \cref{eq:reparam}. Therefore, the deployed model contains only standard compact transformer blocks and incurs no additional inference-time overhead.

For the annealed activation schedule, we set the warm-up ratio $\rho_{\mathrm{w}}$ to $0.01$ and the final linear-stabilization ratio $\rho_{\mathrm{l}}$ to $0.2$.

\tabBackboneConfig
\tabPostTrainConfig
\section{Implementation Details}
\label{sec:implementation-details}
A major challenge in comparing structured pruning methods is that prior work often differs in recovery datasets, data scale, and evaluation protocols.
To ensure a controlled comparison, we standardize the recovery data and evaluation framework across all methods whenever possible, as described in \cref{sec:recovery-dataset}.
For evaluation, we use \texttt{lm-eval-harness}~\cite{lmevalharness}.

For reasoning benchmarks, we evaluate on ARC-Challenge and ARC-Easy~\cite{peter2018arc}, BoolQ~\cite{clark2019boolq}, HellaSwag~\cite{zellers2019hellaswag}, MathQA~\cite{amini2019mathqa}, MMLU~\cite{hendrycks2020mmlu}, OpenBookQA~\cite{mihaylov2018openbookqa}, PIQA~\cite{bisk2020piqa}, RACE~\cite{lai2017race}, and Winogrande~\cite{sakaguchi2021winogrande}.
For generation benchmarks, we evaluate on CoQA~\cite{reddy2019coqa}, GSM8K~\cite{cobbe2021gsm8k}, and TriviaQA~\cite{joshi2017triviaqa}.
CoQA is measured using F1 score in the zero-shot setting, while GSM8K and TriviaQA are measured by exact match in the 8-shot and 5-shot settings, respectively, following \citet{lmevalharness}.

\subsection{Recovery Dataset}
\label{sec:recovery-dataset}
We use the \texttt{FineWeb-Edu} dataset~\cite{lozhkov2024finewebedu} provided by HuggingFaceFW.
Specifically, we use the \texttt{sample-10BT} split and draw 120,000 training samples and 4,000 test samples with a fixed random seed of 42.
We tokenize all samples with a sequence length of 1024 for memory efficiency.
For pruning-criterion calibration, we randomly select 50 samples from the same training split.
We use this dataset configuration consistently across all experiments, regardless of backbone or method, to ensure a fair comparison.

\subsection{Training Configurations}
We follow each baseline method's official training configuration as closely as possible.
To ensure a fair comparison while obtaining strong performance, we use the official settings as the starting point and conduct a limited hyperparameter search around them for each method and backbone.
Unless otherwise specified, all recovery methods are trained on the same recovery dataset and evaluated using the same benchmark pipeline.

For LaCo, the original recovery stage uses full fine-tuning, which is not feasible on an RTX 3090 GPU with 24GB memory in our setting.
We therefore use LoRA as the recovery module for LaCo.
ShortGPT performs pruning based on block influence and uses LoRA for recovery.
For Streamline, we find that the official training schedule uses too few epochs to achieve stable recovery performance in our setting.
Thus, we increase the number of epochs so that its training time is comparable to that of the other methods.

\Cref{tab:post-train-config,tab:backbone-config} provide the hyperparameters and architectural configurations used for recovery.
We apply the AdamW optimizer with $(\beta_1=0.9, \beta_2=0.95)$ in all experiments.
In \cref{tab:post-train-config}, LoRA, AdaLoRA, RankAdaptor, and RestoreLCC share the same hyperparameters and are collectively referred to as recovery methods.

\tabStreamlineReproduce
\tabConfidenceInterval
\subsection{Reproduction Verification}
\label{sec:reproduction-verification}
Prior pruning studies often use different training and evaluation protocols, making direct comparison difficult. We standardize these factors in our main experiments and further verify the fidelity of our baseline reproduction by evaluating our reproduced Streamline with the original Streamline evaluation framework.
In \cref{tab:streamline-reproduce}, the reproduced results closely match the Streamline values reported in the original paper.
This supports the reliability of our reproduced baselines.

\subsection{Confidence Intervals}
\label{sec:confidence-intervals}
To improve the reliability of our experiments, \cref{tab:confidence-interval} reports the 95\% confidence-interval half-widths computed over three random seeds (0, 26, 42) for pruned LLaMA3-3B on both reasoning and generation benchmarks. Each interval is computed as $t_{0.975,n-1} \cdot s / \sqrt{n}$, where $n=3$ and $s$ is the sample standard deviation across seeds. Thus, an entry $c$ corresponds to a confidence interval of $\pm c$.

\tabRuntime
\tabSameEpoch
\tabEndToEndCost

\subsection{Recovery Training Fairness}
\label{sec:runtime-fairness}

\paragraph{Wall-clock fairness}
\ours introduces substantially more trainable parameters than conventional recovery methods, increasing recovery capacity while raising potential concerns about optimization cost.
To address this, we use method-specific recovery epochs to keep wall-clock training time as comparable as possible across methods, rather than equalizing training tokens.

\Cref{tab:runtime} reports trainable parameters, recovery epochs, and wall-clock time for recovering LLaMA3-3B after 25\% pruning on a single RTX 3090 GPU with 24GB memory.
Here, `LoRA-like' denotes LoRA-based methods, including LoRA, LaCo, and ShortGPT.
Despite having the most trainable parameters, \ours achieves wall-clock recovery time comparable to or lower than existing baselines.
Streamline$^{\ast}$ and \ours{}$^{\ast}$ use cached features, whose precomputation takes less than one hour; even including this overhead, their total recovery time remains comparable.
These results show that the gains of \ours are not due to a substantially larger wall-clock training budget.

\paragraph{Training budget fairness}
Beyond the wall-clock comparison, we also evaluate all methods under the same 10-epoch recovery schedule to control for the number of optimization epochs and recovery samples seen during training. As shown in \cref{tab:same-epoch}, \ours still outperforms the baselines when epoch count is fixed across methods. This indicates that the advantage of \ours is not simply due to longer training or exposure to more recovery data. Together with the wall-clock analysis, these results show that \ours remains effective under multiple fair training-budget comparisons.

\subsection{Training-cost Quantification}
We quantify \ours's cost from two perspectives: end-to-end cost relative to LoRA and intrinsic overhead over an otherwise identical Plain recovery on LLaMA3-3B (L3-3B), Qwen3-14B (Q3-14B), and Qwen3-30B-A3B (Q3-30B). `Time' denotes wall-clock time multiplied by the number of GPUs.

\paragraph{End-to-end cost relative to LoRA}
\ours trains only localized recovery blocks, and reuses cached frozen-prefix activations, avoiding repeated full-model forward and backward passes. Although \ours is trained for 20 epochs versus 10 for LoRA, it requires fewer GPU-hours because optimization is limited to the recovery blocks. We report cache precomputation separately and include it in the end-to-end comparison.
As shown in \cref{tab:app-end-to-end-cost}, LoRA requires eight RTX 3090 GPUs on the two larger backbones, while \ours recovery runs on one GPU. Including cache precomputation, \ours is 6.1$\times$ cheaper on Qwen3-14B and 34$\times$ cheaper on Qwen3-30B-A3B.

\paragraph{Intrinsic overhead of the overcomplete parameterization}
We isolate \ours's additional cost by comparing it with Plain recovery. Plain recovery directly fine-tunes the same localized recovery blocks without the auxiliary $\W_i$ and $\D_i$ components or annealed activation, trained for the same 20 epochs. As shown in \cref{tab:app-intrinsic-overhead}, overcomplete parameterization increases recovery time by 2.4--6.6$\times$ over Plain recovery, while even the largest run requires only 12.6 single-GPU hours. Qwen3-30B-A3B shows a smaller increase in trainable parameters because each expert has a narrower intermediate dimension than dense models.

\section{Generality}
\label{sec:app-generality}

\subsection{OverRep in Channel-wise Pruning}
We extend \ours to channel-wise pruning using the same recovery components. Channel-wise pruning reduces the intermediate width of a block, so we adapt the shapes of $\W_i$ and $\D_i$ while retaining the training formulation in \cref{eq:overcomplete-projection-with-va} and the merge rule in \cref{eq:reparam}. Let $d_{\mathrm{mid}}$ denote the intermediate width of a self-attention or feed-forward block, and let $d_{\mathrm{ref}}<d_{\mathrm{mid}}$ denote the reduced width after pruning. Because the block input and output dimensions remain unchanged, all other network components are unaffected.

\tabIntrinsicOverhead
\paragraph{Expanding projections}
For projections into the intermediate space, $i\in\{q,k,v,\mathit{gate},\mathit{up}\}$, the pretrained weight $\P_i\in\R^{d_{\mathrm{mid}}\times d_i^{\mathrm{in}}}$ and additive branch $\W_i\in\R^{d_{\mathrm{mid}}\times d_i^{\mathrm{in}}}$ retain their original shapes, while the multiplicative factor becomes rectangular, $\D_i\in\R^{d_{\mathrm{ref}}\times d_{\mathrm{mid}}}$. 
After training, the merge rule in \cref{eq:reparam} yields $\hat{\P}_i=\D_i(\P_i+\W_i)\in\R^{d_{\mathrm{ref}}\times d_i^{\mathrm{in}}}$. The merge itself reduces the dimension.

\paragraph{Contracting projections}
Projections out of the intermediate space, $i\in\{o,\mathit{down}\}$, have pretrained weights $\P_i\in\R^{d_i^{\mathrm{out}}\times d_{\mathrm{mid}}}$ and must consume the narrowed hidden state $\h'\in\R^{d_{\mathrm{ref}}}$. We mirror the preceding construction by placing the multiplicative factor on the input side, yielding the merged projection $\hat{\P}_i=(\P_i+\W_i)\D_i\in\R^{d_i^{\mathrm{out}}\times d_{\mathrm{ref}}}$, where $\W_i\in\R^{d_i^{\mathrm{out}}\times d_{\mathrm{mid}}}$ and $\D_i\in\R^{d_{\mathrm{mid}}\times d_{\mathrm{ref}}}$.

\paragraph{Initialization and training}
As in the main setting, $\W_i=\mathbf{0}$. Because $\D_i$ is rectangular, it is initialized as a partial identity, $\D_i=[\,\mathbf{I}_{d_{\mathrm{ref}}}\;\;\mathbf{0}\,]\in\R^{d_{\mathrm{ref}}\times d_{\mathrm{mid}}}$ for expanding projections and its transpose for contracting projections. Training therefore begins from a width-reduced copy of the pretrained block. We keep $\P_i$ frozen and train only $(\W_i,\D_i)$. The annealed activation $\AAct_s$ follows \cref{eq:va,eq:alpha-schedule} without modification, and the recovery objective applies the layer-wise reconstruction loss in \cref{eq:ours} to every pruned block. At deployment, each block folds into standard reduced-width projections without additional parameters or computation.

\tabChannelPruning
\paragraph{Comparison with LLM-Pruner}
We compare against LLM-Pruner~\cite{ma2023llmpruner} on LLaMA3-3B and LLaMA3-8B at approximately 25\% parameter reduction, following its evaluation protocol of nine reasoning tasks, including WSC, without the generation suite. To match the LLM-Pruner setting, these experiments use a smaller recovery budget of 24K \texttt{FineWeb-Edu} samples for 5 epochs. Their absolute results are therefore not directly comparable to those under our main protocol, which uses 120K samples for 20 epochs. As shown in \cref{tab:app-channel-pruning}, \ours improves the average by 5.6 points on LLaMA3-3B and 4.2 points on LLaMA3-8B.

\subsection{OverRep in Various Architectures}

\tabVariousArchitecture
\paragraph{Larger and MoE architectures}
A key property of \ours is that recovery is layer-local: training memory is bounded by the two recovery blocks plus cached features, not by total model size. We scale \ours to Qwen3-30B-A3B-Base, a recent large-scale 30.5B total-parameter MoE model, and Qwen3-14B.
Each expert's FFN projections and the attention projections are expanded with their own $(\W_i, \D_i)$ and merged exactly by \cref{eq:reparam}. The router is left untouched, so expert routing behavior and sparsity are preserved at deployment. As shown in \cref{tab:app-various-architecture}, \ours outperforms LoRA while using 48$\times$ fewer aggregate GPU-hours on Qwen3-30B-A3B and 7.2$\times$ fewer on Qwen3-14B.

\tabNemotron
\paragraph{Hybrid architecture}
\ours acts on linear projections and therefore also extends to the dominant parameterized components of Mamba-style hybrid blocks. We validate this on Nemotron-H-4B~\cite{taghibakhshi2025nemotron} by applying \ours to the attention and MLP sublayers of the recovery blocks and keeping the Mamba-2~\cite{dao2024transformers} mixer frozen. As shown in \cref{tab:app-nemotron}, at 25\% pruning, \ours modestly improves reasoning at less than half the cost, although LoRA remains stronger on generation; \ours-KD improves both metrics. At 50\%, \ours improves reasoning by 3.2 points, and \ours-KD improves generation by 4.6 points. 

\tabPruningCriteria
\subsection{Compatibility with Existing Pruning Criteria}
We apply \ours to both contiguous Streamline masks and non-contiguous ShortGPT block influence masks.
As shown in \cref{tab:app-pruning-criteria}, across all eight settings, \ours improves reasoning in six and generation in seven; the remaining differences are below one point. 
These gains require no criterion-specific tuning, supporting \ours as a broadly compatible recovery framework.
\vfill\null\newpage

\section{Detailed Experimental Results}
\subsection{Reasoning Performance}
\label{sec:detailed-results-reasoning}
\Cref{tab:reasoning-full-llama,tab:reasoning-full-qwen} provide detailed results for reasoning tasks.

\subsection{Generation Performance}
\label{sec:detailed-results-generation}
\Cref{tab:generation-full-llama,tab:generation-full-qwen} provide generation results.

\tabReasoningFullLlama
\tabReasoningFullQwen

\tabGenerationFullLlama
\tabGenerationFullQwen

\end{document}